\documentclass[twoside]{article}

\usepackage[american]{babel}

\usepackage{mathtools}                           %
\usepackage{tikz}                                %
\usepackage[utf8]{inputenc}                      
\usepackage[round,authoryear]{natbib}  
\usepackage[T1]{fontenc}                         
\usepackage{hyperref}                            
\usepackage{url}                                 
\usepackage{booktabs}                            
\usepackage{amsthm,amsmath,amssymb,amsfonts}     
\usepackage[sans]{dsfont} 
\usepackage{thmtools,thm-restate}                
\usepackage{algorithm,algorithmic}               
\usepackage{mdframed}                            
\usepackage[many]{tcolorbox}                     
\usepackage{tabularx}                            
\usepackage{nicefrac}                            
\usepackage{microtype}                           
\usepackage{xcolor}                              
\usepackage{bm}                                  
\usepackage{graphicx}                            
\usepackage{enumitem}                            
\usepackage{aliascnt}                            
\usepackage{sidecap}                             
\usepackage{makecell}                            
\usepackage{array}                               
\usepackage{subcaption}
\usepackage{lmodern}
\usepackage{inconsolata}

\declaretheoremstyle[
  spaceabove=5pt,
  spacebelow=5pt,
  headfont=\normalfont\bfseries,
  notefont=\mdseries,
  notebraces={(}{)},
  bodyfont=\normalfont,
  postheadspace=1em,
  qed=,
  headpunct={},
]{theoremstyle}

\declaretheorem[name=Theorem,style=theoremstyle,numberwithin=section]{theorem}

\declaretheorem[name=Definition,style=theoremstyle,numberwithin=section]{definition}
\declaretheorem[name=Assumption,style=theoremstyle,numberwithin=section]{assumption}

\declaretheorem[name=Remark,style=theoremstyle,numberwithin=section]{remark}

\DeclareMathOperator*{\argmin}{arg\,min}
\DeclareMathOperator{\tr}{Tr}

\def\eg{\emph{e.g.},~}
\def\ie{\emph{i.e.},~}

\def\equationautorefname~#1\null{(#1)\null}

\definecolor{tab:blue}{RGB}{31, 119, 180}
\definecolor{tab:orange}{RGB}{255, 127, 14}
\definecolor{linkcolor}{RGB}{44, 96, 163}
\hypersetup{
    colorlinks=true,
    citecolor=linkcolor,
    linkcolor=linkcolor,
    urlcolor=linkcolor,
}

\def\R{\mathbb R}

\def\E{\mathbb E}
\def\P{\mathbb P}

\def\Var{\mathrm{Var}}
\def\Cov{\mathrm{Cov}}
\renewcommand\epsilon{\varepsilon}

\def\vc{{\mathbf{c}}}
\def\vd{{\mathbf{d}}}

\def\vt{{\mathbf{t}}}
\def\vu{{\mathbf{u}}}
\def\vv{{\mathbf{v}}}
\def\vw{{\mathbf{w}}}
\def\vx{{\mathbf{x}}}
\def\vy{{\mathbf{y}}}

\def \vomega{{\bm{\omega}}}

\def \vdelta{{\bm{\delta}}}
\def \vmu{{\bm{\mu}}}

\def\mA{{\mathbf{A}}}

\def\mC{{\mathbf{C}}}
\def\mD{{\mathbf{D}}}

\def\mF{{\mathbf{F}}}

\def\mI{{\mathbf{I}}}
\def\mJ{{\mathbf{J}}}
\def\mK{{\mathbf{K}}}

\def\mM{{\mathbf{M}}}

\def\mQ{{\mathbf{Q}}}

\def\mV{{\mathbf{V}}}

\def\mX{{\mathbf{X}}}

\def\mSigma {{\mathbf{\Sigma }}}

\def\R{\mathbb{R}}
\def\P{\mathbb{P}}

\def\E{\mathbb{E}}

\def\eg{\emph{e.g.,}~}

\usepackage[accepted]{aistats2026}

\usepackage{tikz}
\usetikzlibrary{positioning}

\usepackage{amssymb}
\DeclareMathOperator{\diag}{diag}

\usepackage{pifont}

\begin{document}

\runningtitle{High-Dimensional Analysis of Bootstrap Ensemble Classifiers}
\runningauthor{Tiomoko, Cherkaoui, Seddik, Louart, Schnoor, Kégl}

\twocolumn[
    \aistatstitle{High-Dimensional Analysis of Bootstrap Ensemble Classifiers}
    \aistatsauthor{
      Malik Tiomoko \\Huawei Noah’s Ark Lab, \\France
      \And
      Hamza Cherkaoui \\SAMOVAR, Télécom SudParis\\ Institut Polytechnique de Paris \\France
      \And
      Mohamed~E.~A.~Seddik \\Technology Innovation Institute, \\United Arab Emirates \AND
      Cosme Louart \\Chinese University of Hong Kong, \\China
      \And
      Ekkehard Schnoor \\Fraunhofer HHI, Germany, \\ Aalto University \\Finland
      \And
      Bal\'azs K\'egl \\Huawei Noah’s Ark Lab, \\France
    }
    \aistatsaddress{}
]

\begin{abstract}
Bootstrap methods have long been the cornerstone of ensemble learning in machine learning. This paper presents a theoretical analysis of bootstrap techniques applied to the Least Square Support Vector Machine (LSSVM) ensemble in the context of large and growing sample sizes and feature dimensionalities. Using tools from Random Matrix Theory, we investigate the performance of this classifier that aggregates decision functions from multiple weak classifiers, each trained on different subsets of the data. We provide insights into the use of bootstrap methods in high-dimensional settings, enhancing our understanding of their impact. Based on these findings, we propose strategies to select the number of subsets and the regularization parameter that maximize the performance of the LSSVM. Empirical experiments on synthetic and real-world datasets validate our theoretical results.
\end{abstract}

\section{INTRODUCTION}
\label{submission}
Bootstrap methods~\citep{efron1979bootstrap} are a cornerstone of modern statistics and machine learning. By resampling data and aggregating models, ensemble techniques such as bagging~\citep{breiman1996bagging} reduce variance, improve robustness, and enhance generalization~\citep{efron1994introduction}. Despite their empirical success, however, their behavior in high-dimensional regimes remains only partially understood. 

Classical bootstrap analyses assume that the sample size $n$ grows while the dimension $d$ is fixed~\citep{freedman1981bootstrapping}. These results break down when $n$ and $d$ grow jointly—a regime that is now ubiquitous in modern applications. Recent works have shown that bootstrap procedures can become inconsistent or severely biased in high dimensions~\citep{elkaroui2018bootstrap, clarte2024analysis, bellec2024resampling}, raising fundamental questions: \emph{When does bootstrapping remain reliable? How should one choose the resampling strategy and regularization in high dimensions?}
\begin{figure}[t!]
  \centering
  \includegraphics[width=0.45\textwidth]{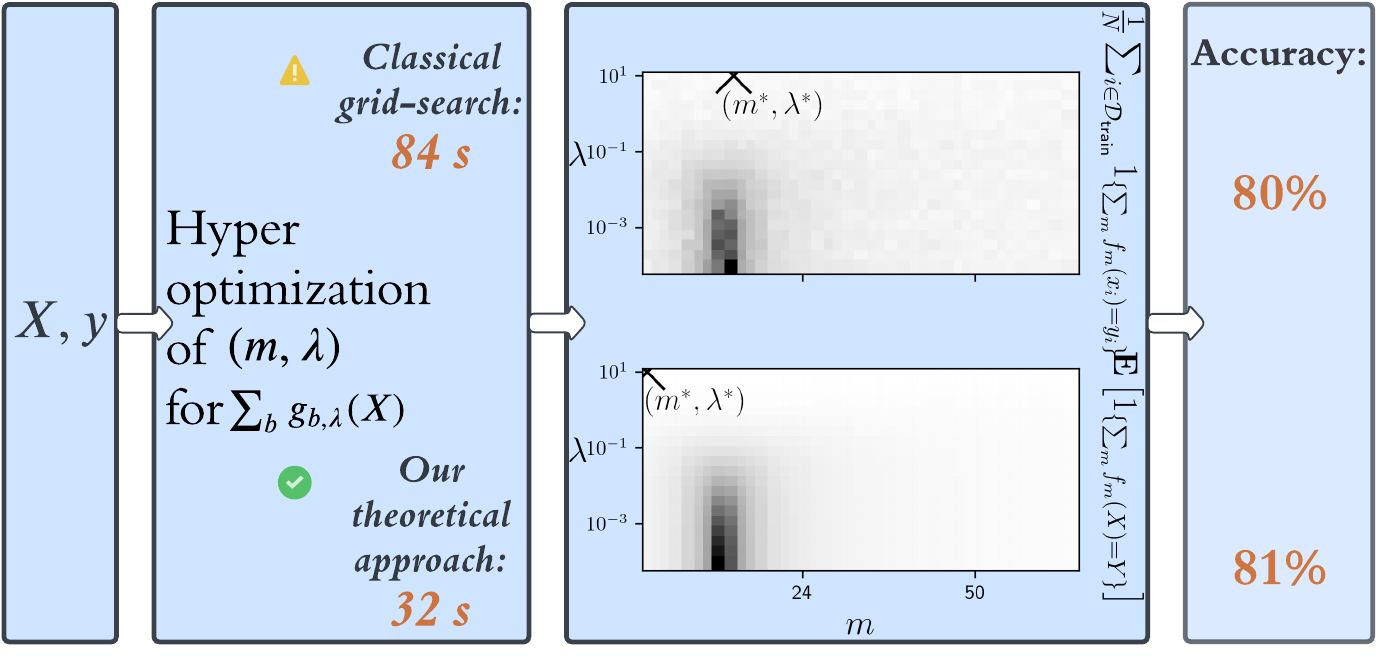}
  \vspace{-3.0mm}
  \caption{From costly grid search to theory-guided tuning. Heatmaps illustrate how hyperparameter choices (regularization and ensemble size) affect the LSSVM decision boundary. Our RMT-based analysis provides closed-form approximations that guide near-optimal settings without exhaustive search.}
  \label{fig:intro_figure}
\end{figure}

We address these questions in the context of \emph{high-dimensional classification ensembles}. In contrast to prior works that primarily focus on regression or rely on Gaussian assumptions, we study bootstrap Least Squares Support Vector Machines (LSSVMs) under a general class of \emph{concentrated random vectors}, which captures structured data distributions commonly encountered in modern learning pipelines. Leveraging tools from \emph{Random Matrix Theory (RMT)}, we derive an explicit and tractable characterization of the ensemble decision score. In particular, we show that it converges to a Gaussian distribution, for which we obtain closed-form expressions of the mean and variance, leading to an explicit approximation of the classification error. These results enable principled tuning of both the regularization parameter and the ensemble size (see~\autoref{fig:intro_figure}).  

\paragraph{Contributions.}
\begin{enumerate}[itemsep=0.0pt, topsep=0.0pt]
    \item \textbf{Explicit asymptotic characterization.} We derive deterministic equivalents for the mean and variance of the bootstrap ensemble decision score in the proportional regime ($n,d \to \infty$), and show that it converges to a Gaussian distribution.
    
    \item \textbf{Analysis of correlated bootstrap ensembles.} Our analysis accounts for dependencies induced by subsampling (without replacement), leading to new expressions that capture the effect of ensemble size and sampling strategy on performance.
    
    \item \textbf{Closed-form classification error.} We obtain an explicit approximation of the misclassification rate as a function of data statistics, regularization, and ensemble size.
    
    \item \textbf{Theory-driven model selection.} We propose a principled method to select hyperparameters without cross-validation, based on the derived asymptotic risk, leading to more stable and interpretable tuning procedures.
\end{enumerate}

\paragraph{Relation to prior work.}
Most high-dimensional analyses of bootstrap methods focus on regression or parameter estimation~\citep{du2023ridge, koriyama2024bagging, bellec2024resampling}, or rely on techniques from statistical mechanics and replica methods, which often yield implicit characterizations. In contrast, we consider classification ensembles under a broader class of data models and derive \emph{explicit} and interpretable expressions for both the decision score and the classification error. This distinction is crucial for enabling practical model selection.  

\paragraph{Takeaway.}
Overall, our results show that bootstrap ensembles in high dimensions can be understood as predictable systems with tractable behavior, rather than purely heuristic tools. This perspective provides both theoretical insight and practical benefits for designing and tuning ensemble classifiers.

The remainder of the paper is structured as follows: 
\autoref{sec:related} reviews related work, 
\autoref{sec:models} introduces the bootstrap LSSVM framework and statistical model, 
\autoref{sec:th} presents our theoretical results, 
\autoref{sec:discussion} discusses their implications, 
\autoref{sec:exp} reports empirical validation, 
and \autoref{sec:ccl} concludes.   

\paragraph{Notation.} 
Matrices are denoted by capital letters, e.g., $\mX \in \R^{d \times n}$ and $\mI_d$. 
Vectors are in bold lowercase (e.g., $\vv \in \R^d$), with $\bm{0}_d$ and $\mathds{1}_d$ for all-zeros and all-ones. 
For $\vv \in \R^d$, $\mathcal{D}_\vv$ is the diagonal matrix with $\vv$ on its diagonal, and $\tr(\mA)$ denotes the trace of square matrix $\mA$. 
Scalars appear in plain letters ($a$, $\lambda$). 
Indices $i,j$ denote data samples; $\ell \in \{1,2\}$ denotes class labels.  

We use standard probability notation: $\E[X]$ for expectation, $\Var(X)$ for variance, and $\mSigma_{\vv} = \Cov(\vv)$ for covariance, where $\bar{\vv}$ denotes the mean (or its deterministic equivalent). 
For $p \geq 1$, $\|\cdot\|_p$ denotes the $\ell_p$-norm, and $\|\cdot\|_F$ the Frobenius norm. 
Concatenation is denoted by $[\vv_1,\vv_2]^\top \in \R^{2d}$ (horizontal) and $[\vv_1 | \vv_2] \in \R^{d \times 2}$ (vertical).

\section{RELATED WORK}
\label{sec:related}

Understanding bootstrap methods in high-dimensional regimes requires connecting several lines of work spanning statistics, machine learning, random matrix theory, and statistical mechanics. We organize the literature into four main directions: (i) classical bootstrap and ensemble methods, (ii) high-dimensional failures of resampling, (iii) constructive asymptotic analyses of bagging and subsampling, and (iv) model averaging and risk estimation.

\paragraph{Classical bootstrap and ensemble methods.}
The bootstrap, introduced by \citet{efron1979bootstrap}, was theoretically justified in low-dimensional settings~\citep{freedman1981bootstrapping, singh1981asymptotic}. Ensemble learning methods, including early work in neural networks~\citep{hansen1990neural, perrone1993improving, krogh1995neural}, culminated in bagging~\citep{breiman1996bagging}, which improves prediction by averaging models trained on resampled datasets. These results rely on fixed-dimensional asymptotics and do not extend to modern high-dimensional regimes.

\paragraph{Failures of bootstrap in high dimension.}
When both the sample size $n$ and dimension $d$ grow proportionally, classical bootstrap procedures can break down. \citet{elkaroui2018bootstrap} show that residual and pairs bootstrap can be inconsistent for variance estimation in regression, while \citet{karoui2019spectral} demonstrate failures in spectral statistics. These works highlight that naive resampling is not reliable in high-dimensional settings.

\paragraph{Constructive high-dimensional analyses.}
Recent work has focused on deriving precise asymptotic characterizations of resampling methods. Using tools from random matrix theory, \citet{clarte2024analysis} analyze bootstrap and subsampling for regularized regression and generalized linear models, establishing conditions under which they remain valid. \citet{bellec2024resampling} derive fixed-point equations describing correlations in subsampled estimators, while \citet{koriyama2024bagging} provide exact asymptotics for bagged $M$-estimators. Related works also study subsampled ridge ensembles and random-feature models~\citep{du2023ridge, loureiro2022fluctuations}.

In parallel, a complementary line of work based on statistical mechanics and replica methods has provided sharp asymptotic predictions for ensemble methods. Early contributions~\citep{sollich1995learning, krogh1997statistical} established fundamental connections between ensemble learning, overfitting, and regularization effects. More recent works extend these techniques to modern settings, analyzing subsampling, under-bagging, and ensemble methods for classification and variable selection~\citep{takahashi2023underbagging, takahashi2025replica}. These approaches typically yield implicit characterizations of performance through fixed-point equations.

\paragraph{Model averaging and risk estimation.}
Ensembling can be interpreted as a form of implicit regularization. \citet{lejeune2019ols} show that averaging least-squares estimators over subsets induces a ridge-like effect, while \citet{ando2023modelavg} analyze optimal weighting in high-dimensional model averaging. For hyperparameter tuning, cross-validation is widely used, but its validity is not guaranteed: \citet{bellec2023cgcv} show inconsistency of generalized cross-validation in some ensemble settings, whereas \citet{du2023ridge} establish consistency results under proportional asymptotics.

\paragraph{Contrast with our work.}
Most prior analyses focus on regression, spectral properties, or inference, and often rely on Gaussian assumptions or yield implicit characterizations through fixed-point equations. In contrast, we study \emph{classification ensembles} based on LSSVMs under a general class of concentrated random vectors. Our approach provides \emph{explicit and interpretable} expressions for the mean and variance of the decision score, leading to a closed-form approximation of the classification error. This explicit characterization enables practical and stable model selection, distinguishing our contribution from prior work.

A structured comparison with representative works is provided in Table~\ref{tab:related_works} (Appendix~\ref{sec:related_extended}).

\section{PRELIMINARIES}
\label{sec:models}

\subsection{Least Squares SVM (LSSVM)}

We consider a binary classification problem with data $\{(\vx_i, y_i)\}_{i=1}^n$, where $\vx_i \in \mathbb{R}^d$ are the feature vectors, and $y_i \in \{-1, +1\}$ are the corresponding class labels. The Least Squares Support Vector Machine (LSSVM) estimator $\vomega^\star$ solves the following regularized least-squares optimization problem:
\[
\min_{\vomega \in \mathbb{R}^d} \; \frac{1}{n} \sum_{i=1}^n \left( y_i - \vx_i^\top \vomega \right)^2 + \frac{\lambda}{2} \|\vomega\|_2^2,
\]
where $\lambda > 0$ is the regularization parameter that controls the trade-off between fitting the data and avoiding overfitting. The closed-form solution to this optimization problem is given by:
\begin{equation}
{\vomega^\star}^\top = \frac{1}{n} \vy^\top \mX^\top \left( \frac{1}{n} \mX \mX^\top + \lambda \mI_d \right)^{-1}, \quad \vomega^\star \in \mathbb{R}^d,
\label{eq:LSSVM_solution_general}
\end{equation}
where $\mX = [\vx_1, \dots, \vx_n] \in \mathbb{R}^{d \times n}$ is the data matrix, and $\vy = (y_1, \dots, y_n)^\top$ is the vector of class labels. The classification decision for a test point $\vx$ is made based on the sign of the decision score $g(\vx) = \vx^\top \vomega^\star$, i.e., $\mathrm{sign}(g(\vx))$.

\paragraph{Role of Regularization.} The regularization parameter $\lambda$ plays a crucial role in controlling the complexity of the model. A smaller value of $\lambda$ leads to a model that is more flexible and can overfit the data, while a larger value increases the penalty for large values of $\vomega$, thus leading to a simpler model that may underfit. In high-dimensional settings where the number of features $d$ is comparable to or larger than the number of samples $n$, proper tuning of $\lambda$ is essential to avoid both overfitting and underfitting.

\subsection{Bootstrap LSSVM Ensembles}

To stabilize performance and reduce variance, we construct an ensemble of $m$ bootstrapped classifiers. From the training set, $m$ resampled subsets $\{(\mX_i, \vy_i)\}_{i=1}^m$ are drawn, each of size $n_i$. The $i$-th classifier is trained using the LSSVM formulation and yields a weight vector:
\[
\vomega_i^\top = \frac{1}{n_i} \vy_i^\top \mX_i^\top \left( \frac{1}{n_i} \mX_i \mX_i^\top + \lambda_i \mI_d \right)^{-1},
\]
where $\lambda_i$ is the regularization parameter for the $i$-th bootstrap subset. The corresponding decision score is $g_i(\vx) = \vx^\top \vomega_i$.

The ensemble average is computed as:
\[
g(\vx) = \frac{1}{m} \sum_{i=1}^m g_i(\vx),
\]
and the test point $\vx$ is classified as $\mathcal{C}_2$ if $g(\vx) > 0$ and as $\mathcal{C}_1$ otherwise. This averaging process helps reduce variance by mitigating the impact of fluctuations across bootstrap replicates. In our theoretical analysis, the decision score $g(\vx)$ plays a central role, as its mean and variance fully characterize the asymptotic classification error.

\subsection{High-Dimensional Regime and Data Assumptions}

We work in a proportional-growth setting, where both the sample size $n$ and feature dimension $d$ grow simultaneously, but their ratio remains constant. This allows the application of tools from Random Matrix Theory (RMT) to derive asymptotic results for the bootstrap ensemble classifiers. This setting is particularly relevant in modern machine learning applications where both $n$ and $d$ are large.

\begin{assumption}[Large $n$, large $d$]
\label{ass:asymptotics_modern}
As both $n$ and $d$ tend to infinity, their ratio $\frac{d}{n}$ converges to a constant $c_0 \in (0, \infty)$. Each subset size $n_i$ scales as $n_i/n \to c_i > 0$, while the number of bootstrap replicates $m = \mathcal{O}(1)$. Furthermore, the class proportions converge to constants $\mathbf{c} = (c_1, c_2)$.
\end{assumption}

\begin{assumption}[Concentrated random vectors]
\label{ass:concentrated_random_vector}
Each column of $\mX$ is an independent random vector with mean $\vmu_\ell$ and covariance $\mSigma_\ell$ conditional on its class $\mathcal{C}_\ell$. Furthermore, we assume that $\vx$ satisfies an exponential concentration inequality. Specifically, for any 1-Lipschitz function $\varphi: \mathbb{R}^d \to \mathbb{R}$,
\[
\P\left( |\varphi(\vx) - \E[\varphi(\vx)]| \ge t \right) \le C e^{-(t/\sigma)^q}, \qquad \forall t > 0,
\]
for some constants $C > 0$, $q > 0$, and dimension-independent $\sigma$. We denote this property as $\vx \propto \mathcal{E}_q(\sigma \,|\, \|\cdot\|_2)$.
\end{assumption}

\paragraph{Interpretation.}
Exponential concentration implies that high-dimensional data behaves almost deterministically with respect to any Lipschitz-continuous function. In other words, fluctuations around the mean are exponentially unlikely. This assumption is satisfied by many natural distributions, such as (a) isotropic Gaussian vectors ($q = 2$), (b) uniformly distributed vectors on the sphere, and (c) any Lipschitz transformations of these (e.g., features produced by deep generative models~\citep{seddik2020random}).

This property allows for the concentration of sample averages and quadratic forms, which is key for deriving deterministic equivalents for the decision score $g(\vx)$.

\medskip
For later use, we define the class means $\vmu_\ell = \E[\vx \mid \vx \in \mathcal{C}_\ell]$, covariances $\mSigma_\ell = \Cov(\vx \mid \vx \in \mathcal{C}_\ell)$, the mean matrix $\mM = [\vmu_1 | \vmu_2] \in \mathbb{R}^{d \times 2}$, and generalized covariances $\mC_\ell = \mSigma_\ell + \vmu_\ell \vmu_\ell^\top$.

\begin{figure}[ht]
\centering
\begin{tikzpicture}[
  box/.style={
    draw, rounded corners, fill=blue!10,
    text width=3.8cm, align=center,
    minimum height=0.8cm, font=\small
  },
  arrow/.style={->, thick, >=stealth}
]

\node[box] (data) {Data $\{\vx_i, y_i\}_{i=1}^n$ \\ High-dimensional features $\vx_i \in \mathbb{R}^d$};
\node[box, below right=0.25
cm of data] (bootstrap) {Bootstrap $m$ subsets \\ $\{(\mX_i, \vy_i)\}_{i=1}^m$};
\node[box, below left=0.25cm of bootstrap] (lssvm) {Train LSSVM on each subset \\ $\vomega_i = \text{argmin} (\dots)$};
\node[box, below right=0.25cm of lssvm] (scores) {Compute decision scores \\ $g_i(\vx) = \vx^\top \vomega_i$};
\node[box, below left=0.25cm of scores] (ensemble) {Average ensemble \\ $g(\vx) = \frac{1}{m} \sum_i g_i(\vx)$};
\node[box, below right=0.25cm of ensemble] (theory) {Theoretical Analysis \\ Mean \& Variance of $g(\vx)$ \\ Classification error $\epsilon(m,\lambda)$};

\draw[arrow] (data) -- (bootstrap);
\draw[arrow] (bootstrap) -- (lssvm);
\draw[arrow] (lssvm) -- (scores);
\draw[arrow] (scores) -- (ensemble);
\draw[arrow] (ensemble) -- (theory);

\end{tikzpicture}
\caption{Flowchart of the bootstrap LSSVM ensemble framework. Data is bootstrapped, LSSVMs are trained on each subset, decision scores are computed and averaged, and theoretical analysis predicts classification error.}
\label{fig:framework_flowchart}
\end{figure}
\section{ASYMPTOTIC DISTRIBUTION OF THE DECISION SCORE}
\label{sec:th}

In this section, we characterize the behavior of the ensemble decision score $g(\vx)$ in the high-dimensional regime $n,d\to\infty$ with $d/n \to c_0\in(0,\infty)$.  
We prove that $g(\vx)$ converges in distribution to a Gaussian random variable whose mean and variance can be expressed in closed form.  
This result provides the foundation for computing the asymptotic classification error in terms of the regularization parameter $\lambda$ and ensemble size $m$.

\subsection{Deterministic Equivalents: Key Quantities}

The central tool is a deterministic equivalent of the resolvent matrix, which captures the effect of data covariance and class proportions.  
We introduce the following fixed-point system:
\begin{align}
\bar{\mQ} &= \left( \sum_{\ell=1}^2 \frac{n_\ell}{n}\frac{\mC_\ell}{1+\delta_\ell} + \lambda \mI_d \right)^{-1}, 
\label{eq:Qbar_fixed_point}\\
\delta_\ell &= \frac{1}{n}\tr(\mC_\ell \bar{\mQ}), \qquad \ell\in\{1,2\},
\end{align}
where $\mC_\ell=\mSigma_\ell+\vmu_\ell\vmu_\ell^\top$ are the generalized covariance matrices.  

Intuitively, $\bar{\mQ}$ plays the role of an “effective inverse covariance” adjusted by both regularization ($\lambda$) and bootstrap resampling.  
The correction terms $\delta_\ell$ encode the high-dimensional shrinkage that vanishes in the classical $d\ll n$ regime but remains essential when $d\sim n$.  

To describe fluctuations of $g(\vx)$, we also define $\mK_\ell$, a covariance proxy matrix:
\[
\mK_\ell = \bar{\mQ}\mSigma_\ell \bar{\mQ} 
+ \sum_{\ell'=1}^2 \frac{n_{\ell'}}{n}\,\frac{d^{(\ell)}_{\ell'}}{(1+\delta_{\ell'})^2} 
\, \bar{\mQ}\mC_{\ell'}\bar{\mQ},
\]
where the coefficients $d^{(\ell)}_{\ell'}$ are obtained by solving
\[
\vd^{(\ell)} = \bigl(\mI_2 - \tilde{\mV}\tilde{\mA}\bigr)^{-1}\bar{\vt}^{(\ell)}.
\]
Here,
\begin{align*}
\bar{\vt}^{(\ell)} &= \tfrac{1}{n}\bigl[\tr(\mSigma_\ell \bar{\mQ}\mSigma_1 \bar{\mQ}), \;
\tr(\mSigma_\ell \bar{\mQ}\mSigma_2 \bar{\mQ})\bigr]^\top,\\
\tilde{\mV} &= \tfrac{1}{n}\!\begin{pmatrix}
\tr(\mSigma_1 \bar{\mQ}\mSigma_1 \bar{\mQ}) & \tr(\mSigma_1 \bar{\mQ}\mSigma_2 \bar{\mQ}) \\
\tr(\mSigma_2 \bar{\mQ}\mSigma_1 \bar{\mQ}) & \tr(\mSigma_2 \bar{\mQ}\mSigma_2 \bar{\mQ})
\end{pmatrix}, \\
\tilde{\mA} &= \tfrac{1}{n}\diag\!\left(\tfrac{n_1}{(1+\delta_1)^2},\,\tfrac{n_2}{(1+\delta_2)^2}\right).
\end{align*}

Finally, we define the correction vector
\[
\vdelta' = \bigl(\delta_1',\delta_2'\bigr) 
= \left(\tfrac{1}{n}\tr(\mSigma_1 \mK_\ell),\, \tfrac{1}{n}\tr(\mSigma_2 \mK_\ell)\right),
\]
which will appear in the variance expression.

\subsection{Main Result: Gaussian Limit}

\begin{theorem}[Asymptotic Distribution of the Decision Score]
\label{thm:asymptotic_distribution}
Under Assumptions~ \ref{ass:asymptotics_modern} -- \ref{ass:concentrated_random_vector},  
the ensemble decision score of a test point $\vx\in\mathcal{C}_\ell$ converges in distribution to a Gaussian:
\[
g(\vx) \;\xrightarrow[n,d\to\infty]{}\; \mathcal{N}(\mathfrak{m}_\ell,\sigma_\ell^2),
\]
where the class-specific mean $\mathfrak{m}_\ell$ and variance $\sigma_\ell^2$ are given by deterministic equivalents.
\end{theorem}

\paragraph{Mean.}
\[
\mathfrak{m}_\ell 
= \tilde{\vy}^\top \mD_{\vc}\mD_{\vdelta}\,\mM^\top \bar{\mQ}\,\vmu_\ell,
\]
where $\tilde{\vy}=[-1,1]^\top$, $\mM=[\vmu_1\,|\,\vmu_2]$, and $\mD_{\vc},\mD_{\vdelta}$ are diagonal matrices of class proportions and fixed-point terms.

\paragraph{Variance.}
\begin{align*}
\sigma_\ell^2 &= \tfrac{1}{m}\tilde{\vy}^\top \mV_\ell \tilde{\vy}
+ \tfrac{1}{m}\tilde{\vy}^\top \mD_{\vc}\mD_{\vdelta}\mM^\top \bar{\mV}_\ell \mM \mD_{\vc}\mD_{\vdelta} \tilde{\vy} \\
&\quad - \tfrac{2}{m}\tilde{\vy}^\top \mD_{\vc}\mD_{\vdelta'} \mM^\top \bar{\mQ}\mM \mD_{\vdelta}\mD_{\vc}\tilde{\vy} \\
&\quad + \tfrac{m-1}{m}\tilde{\vy}^\top \mD_{\vc}\mD_{\vdelta}\mM^\top \bar{\mQ}\mSigma_\ell \bar{\mQ}\mM \mD_{\vdelta}\mD_{\vc}\tilde{\vy}.
\end{align*}

The structure of $\sigma_\ell^2$ highlights the trade-off between variance reduction due to averaging ($1/m$ scaling) and additional dependencies across bootstraps.  

\subsection{Theoretical Classification Error}

Having obtained the asymptotic distribution of $g(\vx)$, the misclassification probability can be written in closed form.  

\begin{remark}[Classification Error]
\label{rem:classification_accuracy}
Suppose that for $\vx\in\mathcal{C}_\ell$, the score follows $g(\vx)\sim\mathcal{N}(\mathfrak{m}_\ell,\sigma_\ell^2)$.  
Then the overall error rate is
\[
\epsilon(\eta) = c_1\,\Phi\!\left(\frac{\eta-\mathfrak{m}_1}{\sigma_1}\right)
+ c_2\left[1-\Phi\!\left(\frac{\eta-\mathfrak{m}_2}{\sigma_2}\right)\right],
\]
where $c_\ell$ are class proportions, $\Phi$ is the standard Gaussian CDF, and $\eta$ is the decision threshold.  
The optimal threshold $\eta^\star$ lies between $\mathfrak{m}_1$ and $\mathfrak{m}_2$ and minimizes $\epsilon(\eta)$.  
\end{remark}

\paragraph{Interpretation.}  
The above result formalizes how bootstrap ensembles behave in high dimensions:  
\begin{enumerate}[itemsep=0.0pt, topsep=0.0pt]
\item The \emph{means} $\mathfrak{m}_\ell$ quantify separation between classes.  
\item The \emph{variances} $\sigma_\ell^2$ reflect both intrinsic randomness and ensemble averaging.  
\item The classification error $\epsilon$ depends explicitly on the regularization $\lambda$ and ensemble size $m$, allowing principled tuning without exhaustive grid search.
\end{enumerate}

\paragraph{Generalization to Other Classifiers.} While the results are derived for the LSSVM model, they can be generalized to other linear classifiers, such as logistic regression, under similar assumptions. This extension provides a more comprehensive understanding of ensemble methods in high-dimensional classification.

\paragraph{Exponential Concentration.} The assumption of exponential concentration of the data (Assumption \ref{ass:concentrated_random_vector}) ensures that high-dimensional data behaves deterministically with respect to any Lipschitz-continuous function, which is critical for deriving deterministic equivalents for the decision score. This assumption is satisfied by many common data distributions, including isotropic Gaussian vectors and uniform distributions on the sphere.

\section{DISCUSSION}
\label{sec:discussion}

Our main result, \autoref{thm:asymptotic_distribution}, establishes that the ensemble decision score $g(\vx)$ is asymptotically Gaussian, with explicit formulas for its mean and variance. These two quantities determine the classification error and allow us to understand, in a principled way, how the parameters of the bootstrapped LSSVM interact.

\subsection{Simplified Gaussian Example}

To build intuition, let us specialize to the simplest setting: two Gaussian classes with opposite means $\pm\vmu$ and identity covariance. In this case, the mean of the decision score reduces to
\begin{align*}
\mathfrak{m}_\ell &= (-1)^\ell \frac{\|\vmu\|^2}{1+\lambda+\lambda \delta + \|\vmu\|^2}, \\
\delta &= \frac{\sqrt{(1+\lambda-md/n)^2 + 4\lambda md/n} - (1+\lambda-md/n)}{2\lambda}.
\end{align*}
This expression highlights the key forces at play:
\begin{enumerate}[itemsep=0.0pt, topsep=0.0pt]
    \item The numerator $\|\vmu\|^2$ represents the intrinsic \emph{class separation} (signal).
    \item The denominator contains penalty terms due to \emph{regularization} ($\lambda$), \emph{bootstrap sampling} ($m$), and \emph{dimensionality} ($d$).
\end{enumerate}

Hence, stronger class separation improves performance, while higher regularization, more bootstraps, and higher dimensionality all reduce the effective mean, tightening the decision boundary. These penalties are not purely detrimental: they are the mechanism that prevents overfitting and ensures stability.

\subsection{Variance Contributions}

The variance of the decision score decomposes into four interpretable terms (details in Appendix~\ref{sec:analysis_variance}). While the exact formulas are technical, the main insights can be summarized as follows:
\begin{itemize}[itemsep=0.0pt, topsep=0.0pt]
    \item Regularization–dimension term ($\text{term}_1$): scales as $1/m$, decreases with more classifiers, and stabilizes fluctuations.
    \item Signal-driven terms ($\text{term}_2$): polynomial in $\|\vmu\|^2$, representing how stronger signal also introduces variability. These are suppressed by $\lambda$ and $d$.
    \item Interaction term ($\text{term}_3$): mixes regularization and signal, but decays with larger $m$ and higher $\delta$.
    \item Cross-classifier interaction ($\text{term}_4$): accounts for dependencies across bootstrapped classifiers. For large $m$ it tends to a constant, but vanishes with higher dimensionality.
\end{itemize}

The key qualitative message is:
\begin{enumerate}[itemsep=0.0pt, topsep=0.0pt]
    \item Increasing the ensemble size $m$ reduces variance through $1/m$ averaging, but at the cost of weaker per-classifier signal (since each sees fewer samples).
    \item Regularization $\lambda$ and dimensionality $d$ both act to stabilize fluctuations, but they also shrink the effective separation of the classes.
    \item Stronger signal $\|\vmu\|^2$ boosts both the mean (good) and the variance (bad).
\end{enumerate}

This antagonistic effect of mean versus variance is nothing but the \textbf{bias–variance tradeoff} in a high-dimensional, bootstrapped LSSVM setting.

\subsection{Implications for Classification Error and Model Selection}

From the Gaussian approximation of \autoref{thm:asymptotic_distribution}, the misclassification error depends directly on the ratio
\[
\frac{\mathfrak{m}_1-\mathfrak{m}_2}{\sigma_1+\sigma_2},
\]
which measures signal separation relative to uncertainty. Increasing $m$ initially helps by reducing variance, but after a point the per-classifier weakness dominates, and performance may deteriorate. This explains the non-monotonic dependence of error on $m$ observed in practice.

Similarly, tuning $\lambda$ trades variance reduction for bias: too small $\lambda$ leads to unstable classifiers, too large $\lambda$ shrinks the margin excessively. The dimensionality $d$ has an unavoidable effect: as $d/n$ grows, both mean separation shrinks and variance grows, underscoring the well-known curse of dimensionality.

\paragraph{Practical takeaway.}
Our analysis provides a principled way to tune $(m, \lambda)$ without resorting to exhaustive cross-validation. By computing theoretical means and variances of the decision score, one can estimate classification error in closed form and select parameters that balance the bias–variance tradeoff optimally. For example, one can choose $m$ to maximize variance reduction without excessively weakening the per-classifier signal, and $\lambda$ to prevent overfitting while maintaining a large margin between classes.

\subsection{Summary}
\begin{enumerate}[itemsep=0.0pt, topsep=0.0pt]
    \item The mean of the decision score grows with signal strength $\|\vmu\|^2$ but decreases with regularization $\lambda$, number of bootstraps $m$, and dimension $d$.
    \item The variance shrinks with $m$ and $\lambda$, but grows with $\|\vmu\|^2$ and unfavorable $d/n$ ratios.
    \item The classification error reflects this competition, leading to an optimal regime where moderate $m$ and carefully tuned $\lambda$ outperform extremes.
\end{enumerate}

Overall, the theory unifies classical insights on bias–variance tradeoff with modern high-dimensional random matrix effects, giving a clear roadmap for robust model selection in bootstrapped LSSVMs.
\section{EXPERIMENTS}
\label{sec:exp}

We validate our theoretical results on both synthetic and real datasets. Our objectives are twofold: first, to show that the asymptotic error predictions closely match the observed performance; second, to demonstrate that these predictions can effectively guide hyperparameter selection in practice. All experiments were implemented in Python and run on $30$ \emph{Intel Xeon @ 3.20,GHz} CPUs, with a total runtime of approximately one hour. All code is open source and publicly available at \url{https://github.com/hcherkaoui/optimal_bootstrap}

\subsection{Synthetic Illustration}

We begin with controlled synthetic experiments, where all parameters are explicitly defined. This allows us to isolate the effects of the number of classifiers $m$ and the regularization parameter $\lambda$.

\paragraph{Setup.}  
We consider a binary classification task with $500$ samples per class and feature dimension $d=100$. The two classes, encoded as $\{-1,+1\}$, are drawn from Gaussian distributions with covariance matrices $\bm{\Sigma}_1$ and $\bm{\Sigma}_2$. The class means are symmetric and sparse:
\[
\bm{\mu}_1 = -\bm{\mu}_2 = 0.9 \bm{e}_1,
\]
where $\bm{e}_1 = [1, 0, \dots, 0]^\top \in \mathbb{R}^d$ is the first standard basis vector.  

We build an ensemble of $m$ classifiers $\{g_b\}_{b=1}^m$ (see~\autoref{sec:models}) and measure performance via the empirical classification error:
\[
\tilde{\epsilon}(m,\lambda) = \frac{1}{m}\sum_{i=1}^m \mathbb{I}\left\{\sum_{b=1}^m g_b(x_i) \neq y_i\right\},
\]
with $m \in \{1,\dots,50\}$ and $\lambda \in [10^{-2},10^{-1}]$. We then compare $\tilde{\epsilon}(m,\lambda)$ with the theoretical error $\epsilon(m,\lambda)$ derived in~\autoref{sec:th}. Each configuration was repeated $10$ times and the results were averaged; uncertainty is indicated by transparency.

\subsubsection{Hyperparameter Map Illustration}

First, we set $\bm{\Sigma}_1 = \bm{\Sigma}_2 = \bm{I}_d$, \ie we consider an isotropic covariance.  

\begin{figure}[ht]
    \centering
    \begin{minipage}{0.23\textwidth}
        \centering
        \includegraphics[width=1.1\textwidth]{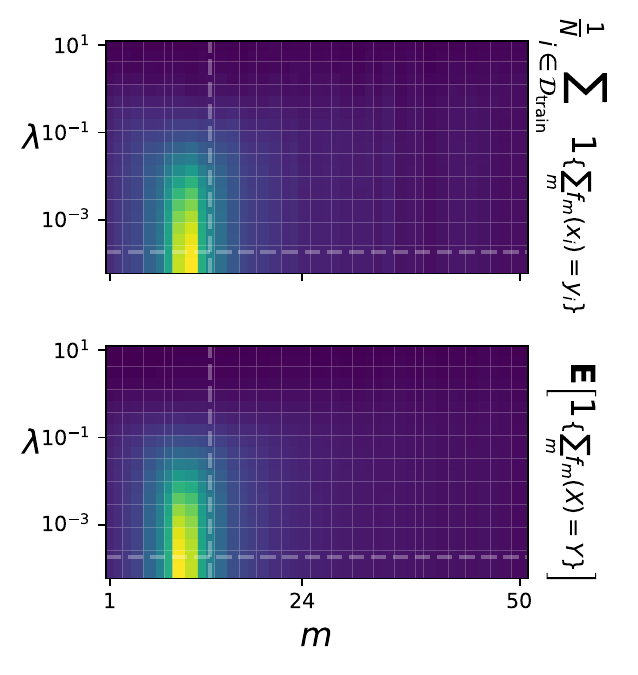}
    \end{minipage}%
    \hfill
    \begin{minipage}{0.23\textwidth}
        \centering
        \includegraphics[width=1.1\textwidth]{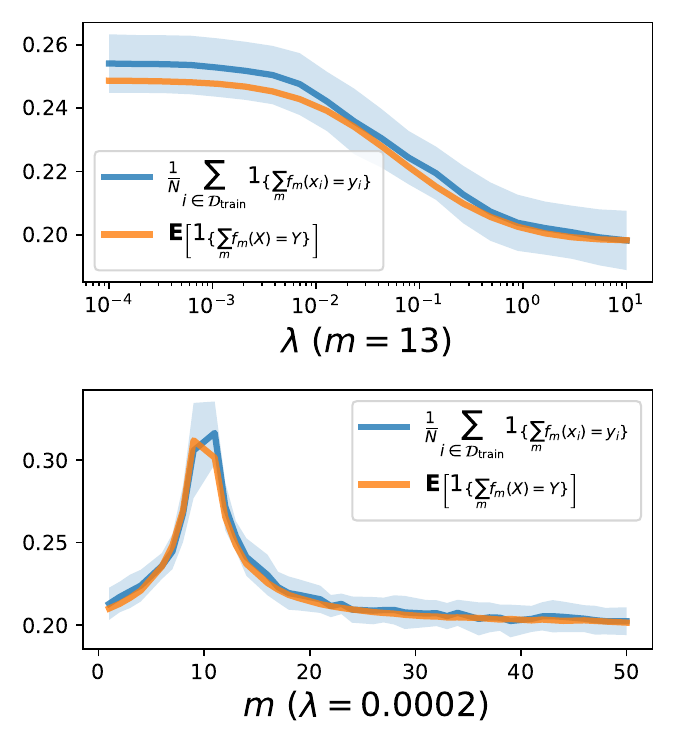}
    \end{minipage}
    \caption{\textbf{(top-left)} Empirical classification error and \textbf{(bottom-left)} theoretical classification error as a function of $m$ and $\lambda$. Darker blue indicates lower error. \textbf{(top-right)} Error as a function of $\lambda$ (for fixed $m=13$). \textbf{(bottom-right)} Error as a function of $m$ (for fixed $\lambda=0.0002$). Empirical errors are in \textcolor{blue}{blue}, theory in \textcolor{orange}{orange}. Shaded regions indicate standard deviation.}
    \label{fig:illustration_map_and_score_evolution}
\end{figure}

As shown in~\autoref{fig:illustration_map_and_score_evolution}, the empirical and theoretical error maps align almost perfectly. Both capture the same low-error and high-error regions, including a clear error peak around $m=10$ when $\lambda=0.0002$. The one-dimensional slices (right panels) confirm this agreement: theoretical curves lie exactly on top of empirical averages, demonstrating the accuracy of our predictions.

\subsubsection{Effect of Covariance Structure}

We next test robustness under different covariance matrices. Alongside the identity covariance, we use a Toeplitz covariance with exponentially decaying entries in its first row.

\begin{figure}[ht]
    \centering
    \begin{minipage}{0.2\textwidth}
        \centering
        \includegraphics[width=\textwidth]{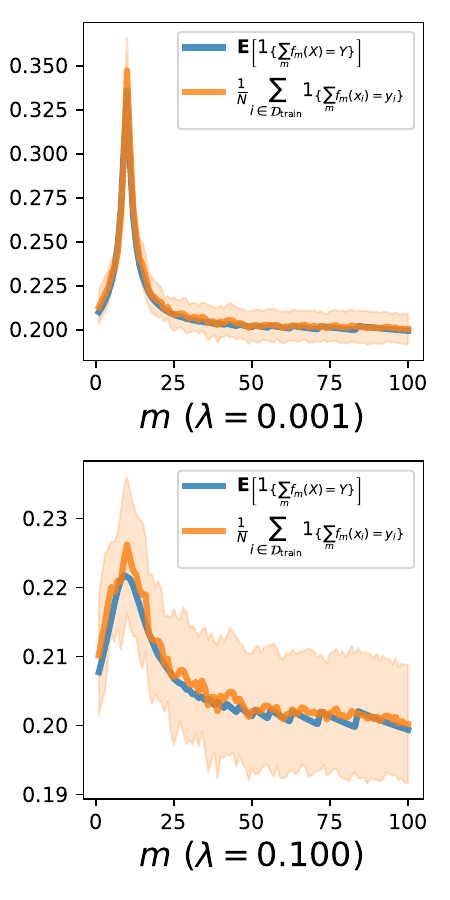}
    \end{minipage}%
    \hfill
    \begin{minipage}{0.2\textwidth}
        \centering
        \includegraphics[width=\textwidth]{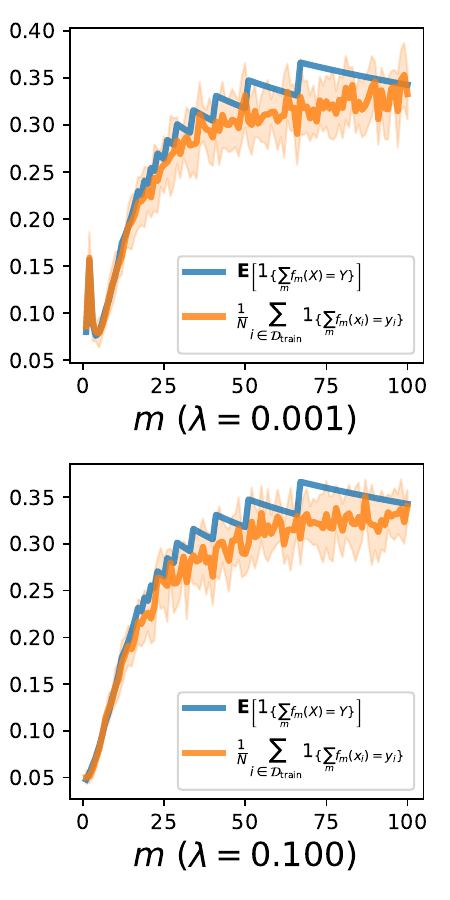}
    \end{minipage}
    \caption{Classification error vs.~$m$ for $\lambda = 0.01$ (\textbf{top}) and $\lambda = 0.1$ (\textbf{bottom}), under identity (\textbf{left}) and Toeplitz (\textbf{right}) covariance. Empirical errors (\textcolor{orange}{orange}) with standard deviation shading vs.~theoretical errors (\textcolor{blue}{blue}).}
    \label{fig:score_evolution}
\end{figure}

In~\autoref{fig:score_evolution}, the theory tracks correctly the empirical error across both covariance settings. Interestingly, qualitative behavior differs: under identity covariance, the error decreases steadily with $m$, whereas under Toeplitz covariance, the lowest error occurs already at $m=1$, and adding classifiers worsens the performance. This highlights how the covariance structure directly influences the optimal ensemble size.

\subsection{Real Data Benchmarks}

We now turn to real-world datasets. Our approach selects hyperparameters by minimizing the theoretical error of~\autoref{thm:asymptotic_distribution}:
\[
(m^*,\lambda^*) = \argmin_{m,\lambda}\, \epsilon(m,\lambda).
\]
We compare our theory-guided hyperoptimization (\texttt{$m^\star$}) against standard hyperparameter selection approaches: grid search (\texttt{Grid}), random search (\texttt{Rand}), a hyperoptimization method (\texttt{HPO} \citep{bergstra2011algorithms,bergstra2022hyperopt}) for different values of $\lambda$.

\subsubsection{Practical Estimation of Moments}

Our framework requires estimates of means and covariance matrices:

\textbf{Means.} We estimate class means via bootstrap averaging: \( \hat{\bm{\mu}}_1 = \mathbb{E}^* [\mathbf{X}_{\mathcal{C}_1}], \quad \hat{\bm{\mu}}_2 = \mathbb{E}^* [\mathbf{X}_{\mathcal{C}_2}]\), which reduces bias and stabilizes estimates in finite samples.  

\textbf{Covariances.} For the estimation of covariances, we use the Ledoit–Wolf shrinkage estimator~\citep{ledoit2004well}: \( \hat{\bm{\Sigma}}_1 = \operatorname{LW}(\mathbf{X}_{\mathcal{C}_1}), \quad \hat{\bm{\Sigma}}_2 = \operatorname{LW}(\mathbf{X}_{\mathcal{C}_2}) \), which is particularly robust in high dimensions.

\paragraph{Remark on Computational Cost.}  An important advantage of our approach is its negligible additional cost. The theoretical classification error depends only on low-dimensional quantities. These quantities are already required for training the base classifiers and can be estimated efficiently (\eg via bootstrap means and Ledoit–Wolf covariance estimators). As a result, computing the theoretical error does not add asymptotic complexity beyond that of training the initial ensemble, making the method computationally efficient and scalable.

\subsubsection{Model Selection on Real Datasets}

We evaluate our method on three representative datasets, chosen to reflect distinct learning contexts where theory and practice may align differently:

\begin{enumerate}[itemsep=1.0pt, topsep=1.5pt]
\item \textit{Pima Indians Diabetes}~\citep{smith1988diabetes}: $n=768$, $d=8$ clinical features, binary label. \emph{Preprocessing:} shuffle and encode targets in ${-1,1}$. This small-dimensional dataset serves as a stress test in a regime where concentration may be weak.

\item \textit{Spambase}~\citep{spambase}: $n=4601$, $d=57$ email features, binary spam label. \emph{Preprocessing:} shuffle and encode targets in $\{-1,1\}$. This dataset provides an intermediate-dimensional benchmark on real tabular data.

\item \textit{Amazon Reviews (Books)}~\citep{blitzer2007domain}: $n=2000$, $d=100$ sentiment embeddings features obtained by applying PCA to the neural network weights, binary polarity task. \emph{Preprocessing:} shuffle, select a balanced subset, PCA, and encode targets in $\{-1,1\}$. This dataset is especially relevant since its deep embeddings align well with our theoretical assumptions.

\item \textit{OpenL3 Audio (4Q)}~\citep{cramer2019openl3}: $n=3167$, $d=100$ audio embeddings features obtained by applying PCA to the neural network weights, binary task obtained by collapsing the original $4$ classes. \emph{Preprocessing:} shuffle, PCA and encode targets in $\{-1,1\}$. This dataset is also especially relevant since its deep embeddings align well with our theoretical assumptions.
\end{enumerate}

\begin{figure}[ht]
  \centering
  \begin{subfigure}{\linewidth}\centering
    \caption{Diabetes}
    \includegraphics[width=0.71\linewidth]{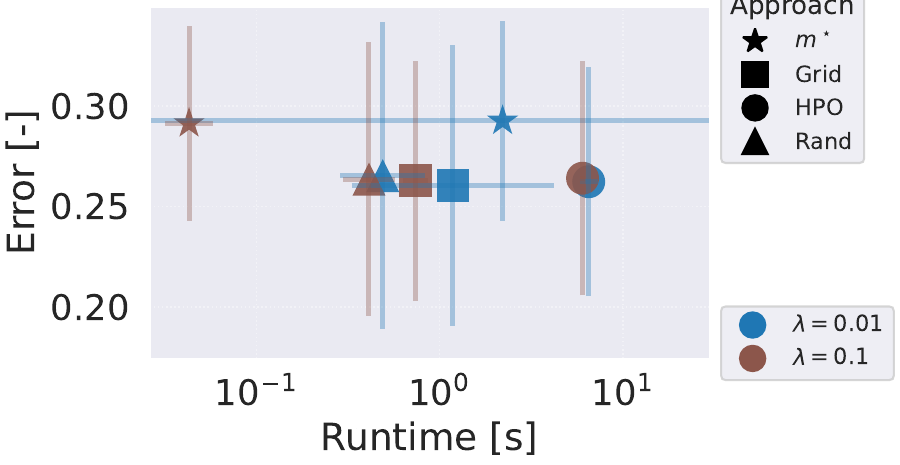}
  \end{subfigure}
  \centering
  \begin{subfigure}{\linewidth}\centering
    \caption{Spams}
    \includegraphics[width=0.71\linewidth]{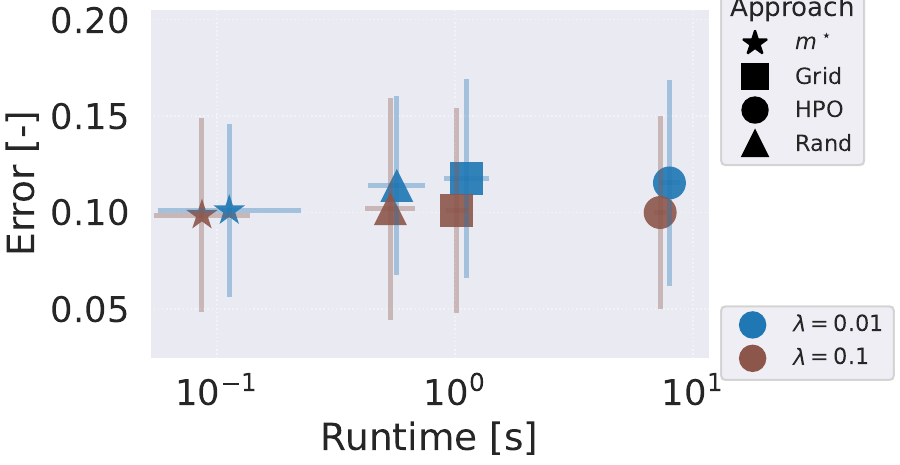}
  \end{subfigure}
  \begin{subfigure}{\linewidth}\centering
    \caption{Amazon review (NN embeddings)}
    \includegraphics[width=0.71\linewidth]{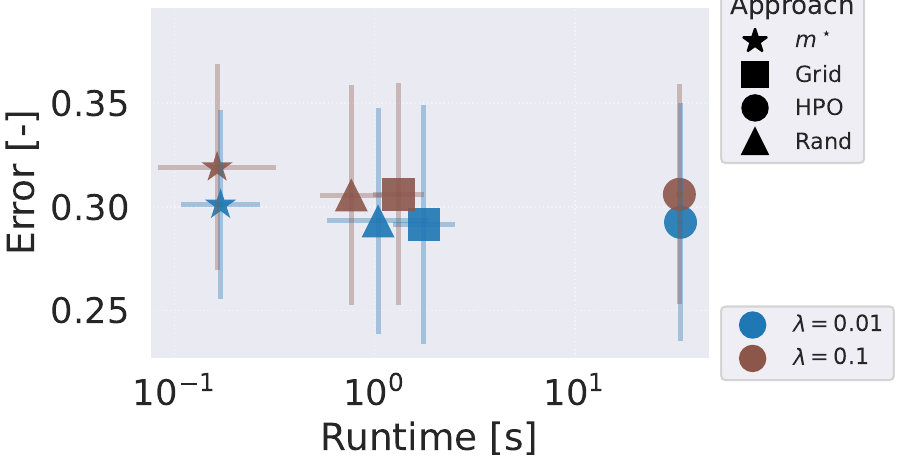}
  \end{subfigure}
  \begin{subfigure}{\linewidth}\centering
    \caption{OpenL3-audio (NN embeddings)}
    \includegraphics[width=0.71\linewidth]{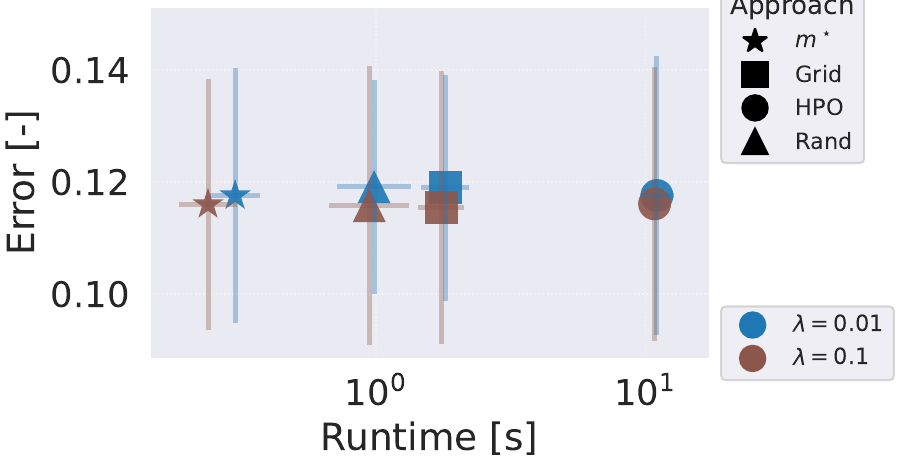}
  \end{subfigure}
  \caption{Performance comparison: test error versus wall-clock runtime across methods and regularization levels $\lambda$.}
  \label{fig:runtime-vs-error-all}
\end{figure}

\paragraph{Discussion.} \autoref{fig:runtime-vs-error-all} shows that our method is consistently competitive in terms of both accuracy and runtime:

\begin{enumerate}[itemsep=1.0pt, topsep=1.5pt]
    \item \textbf{Diabetes:} Our method is slightly weaker in this case, which is expected since the dataset has only $d=8$ features. In such a low-dimensional regime, concentration effects are limited, which reduces the validity of our theoretical assumptions. Including this example is nevertheless important, as it clearly illustrates the limitations of our approach.
    \item \textbf{Spams:} This dataset shows that $d=57$ is already sufficient to place the problem in a regime where our approach is effective: the approximation remains accurate and successfully identifies the hyperparameters yielding the lowest error.
    \item \textbf{Amazon Reviews:} The performance is comparable to that of the best competing methods while requiring less computation time. Since these results are obtained on high-dimensional embeddings, they highlight the practical efficiency gains of our approach in realistic settings.
    \item \textbf{OpenL3:} Our approach again performs similarly to the more costly alternative. It achieves strong predictive performance at a substantially lower computational cost, further confirming the practical relevance of the theory in this setting.
\end{enumerate}

\paragraph{Summary.} Overall, our approach is \emph{fast} and \emph{robust} compared to widely used alternatives. It performs well particularly in the neural embedding regime, which is precisely the class of problems motivating our theoretical framework. On purely low-dimensional tabular data (Diabetes), performance is slightly worse but well-understood, showing that our method is not a universal solution but is most relevant for modern machine learning.

\subsubsection{Limitations: Mixed Feature Types}

Our framework assumes concentrated data vectors, which holds for continuous embeddings but may fail with categorical or mixed features. To test this, we use the \textit{Wisconsin Breast Cancer (Diagnostic)} dataset~\citep{street1993nuclear} ($569$ samples, $30$ features).
\begin{figure}[ht]
    \centering
    \includegraphics[width=0.29\textwidth]{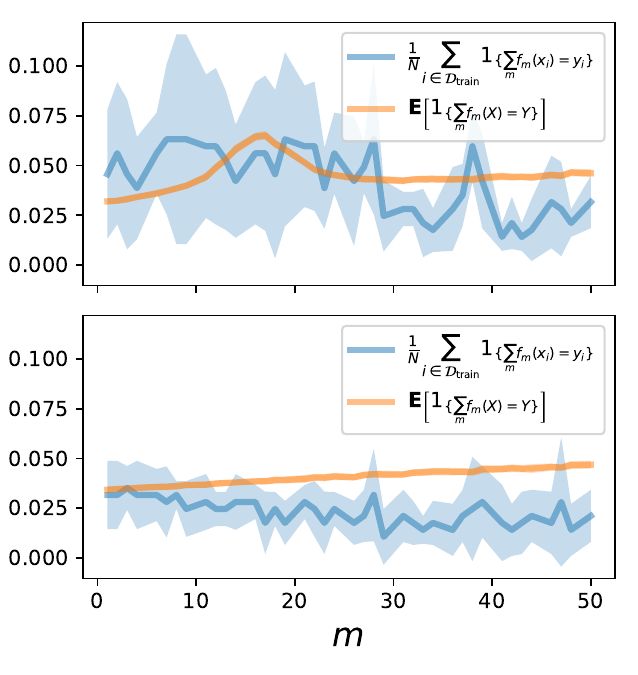}
    \caption{Breast Cancer dataset: empirical error (\textcolor{orange}{orange}) vs.~theory (\textcolor{blue}{blue}).}
    \label{fig:theory_mismatch}
\end{figure}

As shown in \autoref{fig:theory_mismatch}, the empirical error remains essentially flat, whereas the theoretical prediction exhibits a marked increase around $m=17$. This discrepancy is caused by the lack of concentration in the mixed features, which leads to an overestimation of the variance term.

\section{CONCLUSION}
\label{sec:ccl}

We analyzed bootstrap-based ensemble methods for LSSVM using Random Matrix Theory, deriving the asymptotic distribution of the decision score and theoretical classification error. Our results highlight the trade-offs between classifier count, regularization, and data structure. Empirical results confirm the accuracy of our predictions and demonstrate the advantages of a closed-form approach. Future work includes (i) extending to non-linear models (\eg kernelized variants), (ii) multiclass extensions with calibrated outputs, (iii) robustness to class imbalance and covariate shift via reweighting, (iv) compute-aware selection for $(m, \lambda)$ under time/budget constraints, and (v) tighter finite-sample guarantees with improved variance estimators.

\section*{Acknowledgement}
This work was supported by the German Research Foundation (DFG) as research unit DeSBi [KI-FOR 5363] (459422098), by the Research Council of Finland (Decision \#363624) as \textit{A Mathematical Theory of Trustworthy Federated Learning (MATHFUL)}, by the Jane and Aatos Erkko Foundation (Decision \#A835) as \textit{A Mathematical Theory of Federated Learning (TRUST-FELT)}, by the French National Research Agency (ANR) under grant ANR-24-CE40-3341 (project DECATTLON) and by Business Finland as \textit{Forward-Looking AI Governance in Banking \& Insurance (FLAIG)}. This work was granted access to the HPC resources of IDRIS under the allocation 2025-AD011016818, provided by GENCI.

\clearpage
\bibliography{bibliography}

\clearpage
\onecolumn
\aistatstitle{Supplementary Materials for\\ High-Dimensional Analysis of Bootstrap Ensemble Classifiers}


\addtocontents{toc}{\protect\setcounter{tocdepth}{2}}
\renewcommand*\contentsname{\Large Table of Contents}


\section*{Abstract}

This supplementary material complements the main paper by providing (i) additional theoretical details on the bootstrap Least Squares Support Vector Machine (LSSVM) classifier, and (ii) an extensive review of related work. 

On the theoretical side, we revisit in Section~\ref{sec:problem_setup} the bootstrap formulation, where resampled subsets are used to train independent LSSVM classifiers. Section~\ref{sec:theoretical_analysis} develops the Random Matrix Theory (RMT) analysis in detail: we study the distribution of the decision score (\ref{sec:distribution_score}), compute its mean and variance (\ref{sec:mean_computation}, \ref{sec:variance_computation}), and derive the asymptotic classification error (\ref{sec:classification_error}). Additional interpretations and intuitive explanations are provided in Section~\ref{sec:more_insights}. 

On the literature side, Section~\ref{sec:related_extended} expands the related work discussion from the main paper. We provide a comparative table summarizing key contributions across high-dimensional bootstrap, bagging, and ensemble theory, and explicitly contrast them with our work on bootstrap classification ensembles.

\section{Extensive Related Works}
\label{sec:related_extended}

For completeness, we provide here a structured comparison of representative works on bootstrap and ensemble methods in high-dimensional regimes. This complements the discussion in the main paper by offering a compact reference table that situates our contribution relative to prior results in regression, spectral analysis, and inference.

\begin{table*}[t]
\centering
\small
\setlength{\tabcolsep}{6pt}
\renewcommand{\arraystretch}{1.06}
\begin{tabular}{p{4cm} p{1.9cm} p{2.1cm} p{3.2cm} p{3.6cm}}
\toprule
\textbf{Work (ref)} & \textbf{Dim.\ regime} & \textbf{Target} & \textbf{Object / result} & \textbf{Contrast to our work} \\
\midrule
\citet{elkaroui2018bootstrap,karoui2019spectral} & Proportional & Regression / Spectral & Bootstrap and jackknife inconsistency; failures for spectral statistics & Diagnostic negatives; we provide constructive deterministic equivalents for classification scores. \\
\citet{clarte2024analysis} & Proportional & Reg./GLM & Validity conditions for bootstrap, subsampling, and jackknife & Regression focus; we derive classification error maps. \\
\citet{bellec2024resampling} & Proportional & Robust/Logit & Fixed-point correlation equations for subsampling & Regression models; we analyze bootstrap LSSVM ensembles. \\
\citet{koriyama2024bagging} & Proportional & M-estimation & Risk formulas for bagged $M$-estimators & Prediction-risk focus; we target classification error. \\
\citet{bellec2023cgcv} & Proportional & Penalized LS ensembles & Corrected GCV restoring consistency & Risk-estimation focus; we construct analytic error maps. \\
\citet{du2023ridge} & Proportional & Ridge ensembles & Ridge–subsampling equivalences; uniform GCV consistency & Ridge ensembles; we treat LSSVMs with general covariance. \\
\citet{lejeune2019ols} & High-dim & OLS ensembles & Implicit ridge via subsample averaging & Implicit reg.\ in OLS; we quantify ensemble effects on LSSVM scores. \\
\citet{ando2023modelavg} & Proportional & Model averaging & Optimal weights; double-descent phenomena & Model averaging; we derive closed-form classification error maps. \\
\citet{dezeure2016simultaneous} & High-dim sparse & Sparse regression inference & Bootstrap-based simultaneous inference & Sparse inference; our focus is non-sparse classification ensembles. \\
\midrule
\textbf{This work} & \textbf{Proportional, general covariance} & \textbf{Classification (LSSVM)} & \textbf{Deterministic equivalents for score mean/var; closed-form error map} & Constructive tuning of $\lambda$ and $m$ for classification ensembles. \\
\bottomrule
\end{tabular}
\caption{Representative related works organized by regime, target, and results, contrasted with our contribution. ``Proportional'' denotes $n,d \to \infty$ with $d/n \to \gamma \in (0,\infty)$.}
\label{tab:related_works}
\end{table*}

\section{Problem formulation}
\label{sec:problem_setup}
\subsection{The LSSVM Classifier}

We consider a binary classification problem where each observation \( \vx \in \mathbb{R}^d \) is described by \( d \) features, and the output space consists of two classes: \( \mathcal{C}_1 \) and \( \mathcal{C}_2 \). The training set consists of \( n \) labeled examples \( (\mX, \vy) = (\vx_i, y_i)_{i=1}^{n} \), where \( \vx_i \in \mathbb{R}^d \) represents the features of the \(i\)-th sample, and \( y_i \in \{-1, +1\} \) denotes the corresponding class label.

The objective within the Least Squares Support Vector Machine (LSSVM) framework is to find the optimal separating hyperplane that best divides the two classes. This is achieved by solving the following optimization problem:

\begin{equation}
\label{eq:optim}
    \min_{\vomega \in \mathbb{R}^d} \frac{1}{n}\sum_{i=1}^{n} (y_{i} -  \vomega^\top \vx_i)^2 + \frac{\lambda}{2}\|\vomega\|_2^2,
\end{equation}
where \( \lambda > 0 \) is the regularization parameter that controls the trade-off between the goodness of fit to the training data and the smoothness of the solution. The optimization problem in Equation \eqref{eq:optim} is convex and admits a unique closed-form solution for \( \vomega^\star \in \mathbb{R}^d \), which is given by
(for convenience we provide its transpose, as we will consider ${\vomega^\star}^\top \vx$)

\begin{equation}
\label{eq:omegar_star}
    {\vomega^\star}^\top = \frac{1}{n} \vy^\top \mX^\top \left( \frac{\mX \mX^\top}{n} + \lambda \mI_d \right)^{-1},
\end{equation}
where \( \mX \in \mathbb{R}^{d \times n} \) is the data matrix of training samples, and \( \vy \in \mathbb{R}^n \) is the vector of class labels for these samples.
Once the optimal hyperplane is determined, for a new test sample \( \vx \in \mathbb{R}^d \), the classification decision is made based on the sign 
of the decision function \(  \vx^\top \vomega^\star  \), geometrically corresponding to which side of the hyperplane the test point falls into. 
Specifically,  
\[
    \text{Class}(\vx) =
    \begin{cases}
        \mathcal{C}_1 & \text{if } {\vomega^\star}^\top \vx  < 0, \\
        \mathcal{C}_2 & \text{if } {\vomega^\star}^\top \vx  > 0.
    \end{cases}
\]
 
\subsection{Bootstrapping in LSSVM}

Having defined the general framework for Least Squares Support Vector Machines (LSSVM), we now consider the scenario where bootstrapping is applied to improve the robustness of the model. Bootstrapping in the context of LSSVM involves partitioning the entire dataset \( \mX \) into \( m \) subsets. Each subset is then used to train an independent LSSVM classifier. The final decision function is obtained by averaging the decision functions of all individual classifiers, thus combining their predictions for a more stable and accurate outcome.

Let \( \{\mX_i\}_{i=1}^m \) represent the subsets \( m \), where each subset \( \mX_i \in \mathbb{R}^{d \times n_i} \) contains \( n_i \) samples. Each subset \( \mX_i \) is associated with a label vector \( \vy_i \in \{-1, +1\}^{n_i} \).
For simplicity, we assume that all subsets contain the same number of elements, denoted \( n_m \).
Each bootstrapped classifier \( \vomega_i \) is trained on a subset \( \mX_i \) using the LSSVM optimization problem. The decision function for each classifier is given by:

\begin{equation*}
    \vomega_i^\top = \frac{1}{n_i} \vy_i^\top \mX_i^\top \left( \frac{\mX_i \mX_i^\top}{n_i} + \lambda_i \mI_d \right)^{-1},
\end{equation*}
where \( \mX_i \in \mathbb{R}^{d \times n_i} \) and \( \vy_i \in \mathbb{R}^{n_i} \) represent the data and label vectors for the \( i \)-th subset, respectively.

Once individual classifiers are trained, the final classifier is obtained by averaging the decision functions of all bootstrapped classifiers. This ensemble approach helps reduce variance and improve the generalization of the model.
At the heart of this procedure lies the ability to evaluate the performance of the bootstrapped classifier. Specifically, we define the \textit{classification accuracy} as the proportion of correctly classified test samples, and it is given by:

\[
    \text{Accuracy} = \frac{1}{n_{\text{test}}} \sum_{i=1}^{n_{\text{test}}} \mathbb{I}\left(\hat\vy_i \neq \vy_i\right),
\]
where \( \mathbb{I}(\cdot) \) is the indicator function, \( \vx_i \) is the \( i \)-th test sample, \( y_i \) is the true label, and \( \hat\vy_i \) is the predicted label of the bootstrapped classifier.

Our aim in this section is to understand the theoretical behavior of the classification accuracy as a function of the number of bootstrapped classifiers \( m \), the regularization parameter \( \lambda_i \), and other parameters in the model. Specifically, we seek to derive insights into how the number of bootstrapped classifiers impacts the accuracy, and how the averaging of decision functions can help reduce overfitting and improve predictive performance.

\section{Theoretical analysis of bootstrapping in LSSVM}
\label{sec:theoretical_analysis}
In order to predict the theoretical classification error of the bootstrapped LSSVM model, we must first gain a deeper understanding of the statistical behavior of the decision score \( g(\vx) \). The decision score is the key quantity that determines the classification decision for any given test sample. To analyze its effect on classification performance, it is crucial to examine its statistical distribution. This involves computing the first- and second-order moments of \( g(\vx) \), as these will provide the foundation for deriving the theoretical classification error. 

To do this, we will use the following assumptions on the data distribution
\subsection{Data assumptions and growth rate assumption}

\label{app:assumptions}

To perform a theoretical analysis of bootstrapping in LSSVM, we make a few key assumptions about the data. These assumptions will help us derive statistical properties of the decision score and estimate the classification error.

First, we assume a double asymptotic regime in which both the number of samples \( n \) and the dimension \( d \) grow large. This is a common approach in high-dimensional statistics and random matrix theory.

\begin{assumption}[Large \( n \), large \( d \)]
\label{ass:asymptotics}
We assume that \( n > d \), and as both \( n \) and \( d \) increase, the ratio \( d/n \) approaches a constant \( c_0 \in (0,1) \). Furthermore, the fraction of samples in each bootstrap subset, \( n_\ell/n \), tends to a constant \( c_\ell > 0 \) for each subset \( \ell = 1, 2 \). We further assume that the number of subsets does not grow with $n$, 
\ie we assume $m = \mathcal{O}(1)$.
\end{assumption}

Next, we assume that the data matrix \( \mX \) and its columns \( \vx \) (corresponding to individual samples) exhibit a strong concentration, a realistic and mathematically convenient assumption. Intuively, this means that any Lipschitz-continuous observations are tightly clustered around their means with high probability, which also helps to control the behavior of the decision scores based on inner products like \( \vx^\top \vw \). 

\begin{assumption}[Data Concentration]
\label{ass:concentrated_random_vector_appendix}
We assume that the data matrix \( \mX \in \mathbb{R}^{d \times n} \) and its columns \( \vx \in \mathbb{R}^d \) follow a \( q \)-exponential concentration property. Specifically, we assume:
\[
\vx \propto \mathcal{E}_2 (1 \, | \, \mathbb{R}^{d}, \| \cdot \|_2), \qquad \mX \propto \mathcal{E}_2 (1 \, | \, \mathbb{R}^{d \times n}, \| \cdot \|_F),
\]
where \( \mathcal{E}_2 \) denotes \( 2 \)-exponential concentration with respect to the Euclidean and Frobenius norms; see Definition \ref{def:concentrated_random_vector_app}.
\end{assumption}

As a major example, consider \( \vx \sim \mathcal{N}(\bm{0}, \mI) \) for which $\vx \propto \mathcal{E}_2 (1 \, | \, \mathbb{R}^{d}, \| \cdot \|_2)$, \ie the observable diameter is independent of the dimension. Notably, let us also recall that concentration is preserved under Lipschitz-continuous transformations.

 \begin{definition}[$q$-exponential concentration; observable diameter]
 \label{def:concentrated_random_vector_app}
     Let $(\mathcal{X}, \| \, \cdot \, \|_\mathcal{X})$ be a normed vector space and $q > 0$.
     A random vector $\vx \in \mathcal{X}$ is said to be $q$-exponentially concentrated, if 
     for any $1$-Lipschitz continuous (with respect to $\| \, \cdot \, \|_\mathcal{X}$) real-valued function $\varphi: \mathcal{X} \to \R$ there exists
     $C \geq 0$ independent of $\dim(\mathcal{X})$ and $\sigma > 0$ such that 
     \begin{equation*}
         \P \left( |\varphi(\vx) - \E \varphi(\vx)| \geq t \right) \leq C e^{-(t/\sigma)^q} \qquad \forall t > 0.
     \end{equation*}
    This is denoted as $\vx  \propto \mathcal{E}_q (\sigma \, | \, \mathcal{X},  \| \, \cdot \, \|_\mathcal{X})$, where $\sigma$ is called the \textit{observable diameter}.
    If $\sigma$ does not depend on $\dim(\mathcal{X})$, we write
$\vx  \propto \mathcal{E}_q (1 \, | \, \mathcal{X},  \| \, \cdot \, \|_\mathcal{X})$.
 \end{definition}



These assumptions on concentration and growth rates allow us to model the behavior of the decision score and ultimately estimate the classification error of the bootstrapped LSSVM. We also impose the condition that the columns of \(\mX\) are independent. These assumptions ensure the existence of both the mean and covariance for the columns of \(\mX\). For \(\ell \in \{1, 2\}\), we define the following

\[
\vmu_\ell = \mathbb{E}[\vx_i \mid \vx_i \in \mathcal{C}_\ell], \quad \mSigma_\ell = \operatorname{Cov}(\vx_i \mid \vx_i \in \mathcal{C}_\ell)
\]

Additionally, we collect the means of the class distributions into the matrix \(\mM = [\vmu_1 | \vmu_2] \in \mathbb{R}^{d \times 2}\) and we define the generalized covariance as $\mC_\ell = \mSigma_\ell + \vmu_\ell\vmu_\ell^\top$.

\subsection{Distribution of the decision score}
\label{sec:distribution_score}

To demonstrate that the decision score \( g(\vx) = \vw^\top \vx \) follows a Gaussian distribution, we rely on the \textbf{Central Limit Theorem (CLT)} for concentrated random vectors. This result, presented in \citep{klartag2007central} and \citep{fleury2007stability}, shows that the inner product \( \vw^\top \vx \), where \( \vx \) is a random vector satisfying \( q \)-exponential concentration, approaches a Gaussian distribution.

The key conditions for the result are:
\begin{itemize}
    \item \( \vx \) is a random vector with zero mean \( \E[\vx] = \bm{0} \) and identity covariance \( \E[\vx\vx^\top] = \mI_p \).
    \item \( \vx \) satisfies \( q \)-exponential concentration, meaning that the distribution of \( \vx \) is tightly concentrated around its mean, following the concentration property from Definition \ref{def:concentrated_random_vector_app}.
    \item The weight vector \( \vw \) is randomly chosen and independent of \( \vx \).
\end{itemize}

\textbf{Central Limit Theorem}: Under these conditions, for sufficiently large \( p \), the distribution of the inner product \( \vw^\top \vx \) (which is \( g(\vx) \)) is approximately Gaussian. Specifically, it converges to a standard normal distribution \( \mathcal{N}(0, 1) \), with a small deviation bound for the cumulative distribution function \( F(t) \), ensuring that the approximation is accurate up to an error of \( p^{-1/4} \).

This result justifies the assumption that \( g(\vx) \) behaves like a Gaussian random variable, which allows us to compute the theoretical classification error more precisely using numerical methods such as integration.

Furthermore, we define the notion of deterministic equivalent as
\begin{definition}[Deterministic equivalents]
\label{def:deterministic_equivalents}
A deterministic equivalent, say $\bar \mF\in\mathbb{R}^{n\times d}$, of a given random matrix $\mF\in\mathbb{R}^{n\times d}$, denoted $\mF\leftrightarrow \bar{\mF}$, is defined by the fact that, for any deterministic linear functional $f:\mathbb{R}^{n\times d}\to\mathbb{R}$, $f(\mF-\bar \mF)\to 0$ almost surely (for instance, for $\vu,\vv \in \R^d$ of unit norm, $\vu^\top (\mF-\bar{\mF}) \vv\rightarrow 0$ and, for $\mA \in \mathbb{R}^{d\times n}$ deterministic of bounded operator norm, $\frac{1}{n} \operatorname{tr} \mA(\mF-\bar{\mF})\rightarrow 0$).
\end{definition}
Deterministic equivalents are thus particularly suitable to handle bilinear forms involving the random matrix $\mF$.

\subsection{Mean of the Decision Score}
\label{sec:mean_computation}

For a test data point \(\vx \in \mathbb{R}^d\) from class \(\mathcal{C}_\ell\), where \(\ell = 1, 2\), and assuming \(\vx\) is independent from the training data \(\mX \in \mathbb{R}^{d \times n}\), the mean classification score \(g(\vx) = \vw^\top \vx\) can be computed as follows:

\begin{align}
    m_\ell
    = & \frac{1}{m} \sum_{t=1}^m \mathbb{E}_{\mX_t} 
    \left[ \mathbb{E}_\vx \left[ g(\vx) \mid \vx \in \mathcal{C}_\ell \right] \right] 
    \label{first_definition} \\
    = & \frac{1}{m} \sum_{t=1}^m \frac{1}{n} \mathbb{E}_{\mX_t} 
    \left[ \mathbb{E}_\vx \left[ \vy_t^\top \mX_t^\top \left( \frac{1}{n} \mX_t \mX_t^\top + \lambda \mI_d \right)^{-1} \vx \right] \right] 
    \label{expansion_g} \\
    = & \frac{1}{m} \sum_{t=1}^m \frac{1}{n}  \mathbb{E}_{\mX_t}
    \left[ \vy_t^\top \mX_t^\top \left( \frac{1}{n} \mX_t \mX_t^\top + \lambda \mI_d \right)^{-1} \vmu_\ell \right] 
    \label{expectation_test} \\
    = & \frac{1}{m} \sum_{t=1}^m \frac{1}{n}  \mathbb{E}_{\mX_t} \left[ \vy_t^\top \mX_t^\top \mQ_t \vmu_\ell \right] 
    \label{reduction} \\
    = & \frac{1}{n}  \vy_t^\top \mJ \mM_\delta^\top \bar{\mQ} \vmu_\ell
    \label{deterministic_equivalent}
\end{align}

Where \(\bar{\mQ} = \bar{\mQ}(\lambda)\) is the deterministic equivalent of \(\mQ = \left( \frac{1}{n} \mX \mX^\top + \lambda \mI_d \right)^{-1}\), the resolvent of the random matrix \(\frac{1}{n} \mX \mX^\top\). The matrices \(\mJ \in \mathbb{R}^{n \times 2}\) and \(\mM_\delta \in \mathbb{R}^{d \times 2}\) are defined as:

\[
    \mJ = \begin{pmatrix}
        \mathds{1}_{n_1} & \mathbf{0}_{n_1} \\
        \mathbf{0}_{n_2} & \mathds{1}_{n_2}
    \end{pmatrix} \in \mathbb{R}^{n \times 2}, 
    \quad 
    \mM_\delta = \left[ \frac{1}{1 + \delta_1} \vmu_1 \Big| \frac{1}{1 + \delta_2} \vmu_2 \right] \in \mathbb{R}^{d \times 2}
\]

where \(\vdelta = (\delta_1, \delta_2)\). The matrix \(\bar{\mQ}\) is computed iteratively using the fixed-point equation:

\[
    \bar{\mQ} = \left( \sum_{\ell=1}^2 \frac{n_\ell}{n} \frac{\mC_\ell}{1 + \delta_\ell} + \lambda \mI_d \right)^{-1}, 
    \quad 
    \delta_\ell = \frac{1}{n} \tr \left( \mC_\ell \bar{\mQ} \right)
\]

The iterations begin with \(\delta_1 = \delta_2 = 0\) and continue until convergence.

\paragraph{Derivation Details}

- \textbf{From \(\eqref{first_definition}\) to \(\eqref{expansion_g}\):} We apply the definition of the decision score \(g(\vx) = \vy^\top \mX_t^\top \left(\frac{1}{n} \mX_t \mX_t^\top + \lambda \mI_d\right)^{-1} \vx\).
  
- \textbf{From \(\eqref{expansion_g}\) to \(\eqref{expectation_test}\):} We use the independence assumption between the training samples \(\mX_t\) and the test sample \(\vx\). Therefore, the expectation is taken only with respect to the test sample, yielding \(\mathbb{E}[\vx | \vx \in \mathcal{C}_\ell] = \vmu_\ell\).

- \textbf{From \(\eqref{expectation_test}\) to \(\eqref{reduction}\):} This is a rewriting step where we define the resolvent matrix \(\mQ_t = \left(\frac{1}{n} \mX_t \mX_t^\top + \lambda \mI_d\right)^{-1}\).

- \textbf{From \(\eqref{reduction}\) to \(\eqref{deterministic_equivalent}\):} This step involves some technical matrix algebra and tools from Random Matrix Theory, which is explained as follows.

Let \(\mX_{-i}\) represent the matrix \(\mX\) with the \(i\)-th column replaced by \(\mathbf{0}\), such that \(\mX \mX^\top = \mX_{-i} \mX_{-i}^\top + \vx_i \vx_i^\top\). Using the \textit{Sherman-Morrison matrix inversion lemma}, we can express \(\mQ\) as:

\[
    \mQ = \left( \frac{\mX \mX^\top}{n} + \lambda \mI_d \right)^{-1} = \mQ_{-i} - \frac{1}{n} \frac{\mQ_{-i} \vx_i \vx_i^\top \mQ_{-i}}{1 + \frac{1}{n} \vx_i^\top \mQ_{-i} \vx_i}
    \label{eq:Q_i}
\]

Next, we can calculate the interaction between \(\mQ\) and \(\vx_i\):

\[
    \mQ \vx_i = \frac{\mQ_{-i} \vx_i}{1 + \frac{1}{n} \vx_i^\top \mQ_{-i} \vx_i}
    \label{eq:Q_x}
\]

Now, we compute the expectation of the decision score for a given test sample \(\vx_i\):

\[
    \mathbb{E} \left[ \vy_t \vx_i^\top \mQ \vmu_\ell \right] = \frac{1}{n} \sum_{i=1}^n \mathbb{E} \left[ \vy_i \frac{\vx_i^\top \mQ_{-i} \vmu_\ell}{1 + \frac{1}{n} \vx_i^\top \mQ_{-i} \vx_i} \right]
\]

This simplifies to:

\[
    \frac{1}{n} \sum_{i=1}^n \left[ \vy_i \frac{\mathbb{E}[\vx_i]^\top \bar{\mQ} \vmu_\ell}{1 + \delta_{\mathcal{C}(i)}} \right]
\]

where \(\bar{\mQ}\) is the \textit{deterministic equivalent} of \(\mQ\) 
(recall Definition \ref{def:deterministic_equivalents} above)
and provided in \citep{louart2018concentration} as
\[
    \bar{\mQ} = \left( \sum_{\ell=1}^2 \frac{n_\ell}{n} \frac{\mC_\ell}{1 + \delta_\ell} + \lambda \mI_d \right)^{-1}, 
    \qquad \delta_\ell = \frac{1}{n} \tr \left( \mC_\ell \bar{\mQ} \right).
\]

This completes the derivation for the mean of the decision score.

\subsection{Variance of the decision score}
\label{sec:variance_computation}

Next, we compute the variance $\sigma_\ell^2$ of the classification score $g(\vx) = \vx^\top \vw$ for a test dataset $\vx \in \mathcal{C}_\ell$:

\begin{align}
    \sigma_\ell^2 
    = & \Var(g(\vx)) \nonumber \\
    = & \mathbb{E} \left[ g(\vx)^2 \right] - \left( \mathbb{E} [ g(\vx) ] \right)^2 \nonumber \\
    = & \mathbb{E} \left[ \vy^\top \mX^\top \mQ \vx \vx^\top \mQ \mX \vy \right] - \left( \mathbb{E}[g(\vx)] \right)^2 \nonumber \\
    = & \vy^\top \mJ \mM_\delta^\top \mK_\ell \mM_\delta \mJ^\top \vy 
    + \vy^\top \mV_\ell \vy 
    - 2 \vy^\top \mJ \mM_{\vdelta'}^\top \bar{\mQ} \mM_\delta \mJ^\top \vy \label{eq:sigma_ell_final_term}
\end{align}

where $\mK_\ell$, $\mV_\ell$, and other necessary terms are given as follows:

\begin{align}
\tilde{\mV} & = \frac{1}{n} \begin{pmatrix}
         \tr (\mSigma_1 \bar{\mQ} \mSigma_1 \bar{\mQ}) & \tr (\mSigma_1 \bar{\mQ} \mSigma_2 \bar{\mQ}) \\ 
         \tr (\mSigma_2 \bar{\mQ} \mSigma_1 \bar{\mQ}) & \tr (\mSigma_2 \bar{\mQ} \mSigma_2 \bar{\mQ})
    \end{pmatrix} \; \in \R^{2 \times 2}, \\
\tilde{\mA} & = \frac{1}{n} \begin{pmatrix}
          \frac{n_1}{(1 + \delta_1)^2} & 0 \\ 
         0 & \frac{n_2}{(1 + \delta_2)^2}
    \end{pmatrix} \; \in \R^{2 \times 2}, \\
\bar{\vt}^{(\ell)} & = \frac{1}{n} \left[ \tr (\mSigma_\ell \bar{\mQ} \mSigma_1 \bar{\mQ}), \tr (\mSigma_\ell \bar{\mQ} \mSigma_2 \bar{\mQ}) \right]^\top \in \R^{2}, \\
\vd^{(\ell)} & = \left[ d_1^{(\ell)}, d_2^{(\ell)} \right]^\top = \left( \mI_2 - \tilde{\mV} \tilde{\mA} \right)^{-1} \bar{\vt}^{(\ell)} \in \R^{2}, \\
\mK_\ell & = \bar{\mQ} \mSigma_\ell \bar{\mQ} + \sum_{\ell' = 1}^2 \frac{n_{\ell'}}{n} \frac{d_{\ell'}^{(\ell)}}{(1 + \delta_{\ell'})^2} \bar{\mQ} \mC_{\ell'} \bar{\mQ}, \\
\mV_\ell & = \mathcal{D}_{\vv_\ell} \in \R^{n \times n}, \\
\vv_\ell & = \left[ \frac{\tr \left( \mSigma_1 \mK_\ell \right)}{(1 + \delta_1)^2} \mathds{1}_{n_1}, \frac{\tr \left( \mSigma_2 \mK_\ell \right)}{(1 + \delta_2)^2} \mathds{1}_{n_2} \right] \in \R^n
\end{align}

The derivation of the variance follows the same structure as that of the mean, but now the goal is to find the deterministic equivalent of the matrices \( \mQ \mA \mQ \), where \( \mA \) is an arbitrary matrix independent of \( \mQ \) and has bounded norms. The necessary calculations can be carried out in a similar manner to those in \citep{tiomoko2020large} (Section A.2).

\subsection{Derivation of the theoretical error}
\label{sec:classification_error}

After determining the class-specific means and variances of the classification score \( g(\vx) = \vw^\top \vx \) using Section \ref{sec:mean_computation} and \ref{sec:variance_computation}, we can now compute the theoretical classification error, summarized in the following remark.

\vspace{0.15cm}

\begin{remark}[Classification Accuracy]\label{rem:classification_accuracy_general}

For a linear binary classifier \( \vw \in \R^d \), where the classification score \( g(\vx) = \vw^\top \vx \) follows normal distributions for each class, that is, \( \mathcal{N}(m_1, \sigma_1^2) \) for \( \vx \in \mathcal{C}_1 \) and \( \mathcal{N}(m_2, \sigma_2^2) \) for \( \vx \in \mathcal{C}_2 \), the classification error is given by

\[
    \epsilon = c_1 \cdot \P \left( \vx \to \mathcal{C}_2 \mid \vx \in \mathcal{C}_1 \right) + c_2 \cdot \P \left( \vx \to \mathcal{C}_1 \mid \vx \in \mathcal{C}_2 \right),
\]

where \( c_\ell \) denotes the probabilities of the class and \( c_1 = c_2 = \frac{1}{2} \) for balanced classes. The probabilities of misclassification for each class can be expressed as:

\[
    \epsilon = c_1 \cdot \P \left(\vw^\top \vx > \eta \mid \vx \in \mathcal{C}_1 \sim \mathcal{N}(m_1, \sigma_1^2)\right) + c_2 \cdot \P \left(\vw^\top \vx < \eta \mid \vx \in \mathcal{C}_2 \sim \mathcal{N}(m_2, \sigma_2^2)\right),
\]

where \( \eta \) is the decision threshold. These misclassification probabilities are determined by the cumulative distribution functions (CDF) of the normal distributions:
\[
    \epsilon = c_1 \cdot \P \left(X > \eta \mid X \sim \mathcal{N}(m_1, \sigma_1^2)\right) + c_2 \cdot \P \left(Y < \eta \mid Y \sim \mathcal{N}(m_2, \sigma_2^2)\right).
\]

For practical computation, these probabilities can be approximated using numerical integration of the normal distribution densities, given the class-specific means \( m_1, m_2 \) and variances \( \sigma_1^2, \sigma_2^2 \), and the threshold \( \eta \).

The optimal threshold \( \eta \) minimizes the classification error and is typically found between the two means, that is, \( m_1 < \eta < m_2 \), where the densities of the two classes intersect.

\end{remark}

\section{Intuitions from the theoretical analysis}
\label{sec:more_insights}
In this section, we aim to derive several intuitions into the bootstrap LSSVM approach. First we analyze the influence of the mean of the decision score
\subsection{Analysis of the mean of the decision score}
\label{sec:analysis_mean}
We recall that the mean of the  decision score given that the test data is coming from the class $\mathcal{C}_1$ can be written as
\begin{equation*}
    \mathfrak{m}_1 = \frac{1}{n_m}\vy^\top \mJ^\top \mM_\delta^\top\bar{\mQ}\vmu_1
\end{equation*}
Furthermore we will consider first the simple case of mixture of two Gaussian of opposite mean $\pm \vmu$ and identity covariance matrix $\mSigma = \mI_d$. By rewriting the labels $\vy = \mJ\tilde{\vy}$, the mean can be rewritten as
\begin{equation*}
    \mathfrak{m}_1 = c_1 \vmu^\top\bar{\mQ}\vmu
\end{equation*}
Furthermore the deterministic equivalent $\bar{\mQ}$ can be rewritten as
\begin{align*}
    \bar{\mQ} = \kappa \mI_d - \kappa^2 \frac{\vmu\vmu^\top}{(1+\delta) + \kappa\vmu^\top\vmu}
\end{align*}
where $\delta$ is given by the closed form solution
\begin{align*}
    \delta = \frac{\sqrt{(1+\lambda-c_0)^2 + 4\lambda c_0} - (1+\lambda-c_0)}{2\lambda}
\end{align*}
Finally the mean of the decision score can be rewritten as
\begin{align*}
    \mathfrak{m}_1 = \frac{\vmu^\top\vmu}{(1+\lambda+\lambda\delta) + \vmu^\top\vmu}
\end{align*}
As the number of classifiers increases, the ratio $c_0 = p/n$ increases and therefore $\delta$ increases, making the mean to decrease.
The mean only depends on the increasing number of classifier through $c_0$.

The parameter \( m \), given by \( m = \frac{\vmu^\top \vmu}{(1 + \lambda + \lambda \delta) + \vmu^\top \vmu} \), can be interpreted as a normalized measure of the signal-to-noise ratio. Here, \( \vmu^\top \vmu \) represents the squared magnitude of the mean vector \( \vmu \), which measures the separation between the classes. A larger \( \vmu^\top \vmu \) implies stronger discrimination between the two classes, leading to a larger $\mathfrak{m}_1$, while \( \vmu^\top \vmu \to 0 \) results in \( m \to 0 \), indicating no class separation. The term \( 1 + \lambda \) serves as a baseline regularization factor that prevents $\mathfrak{m}_1$ from growing excessively large, even for high \( \vmu^\top \vmu \). The regularization parameter \( \lambda \) controls the trade-off between fitting the data and penalizing model complexity: smaller \( \lambda \) emphasizes data fidelity, whereas larger \( \lambda \) prioritizes simplicity and robustness. The term \( \lambda \delta \), where \( \delta \) increases with \( d/n \), penalizes \( \mathfrak{m}_1 \) to account for the challenges of high-dimensional noise. This highlights the curse of dimensionality, as a larger \( d/n \) reduces \( \mathfrak{m}_1 \), reflecting the difficulty of distinguishing classes in high dimensions. Overall, \( m \) can be expressed as \( \frac{\text{signal}}{\text{signal} + \text{noise}} \), where the signal is \( \vmu^\top \vmu \) (the class separation), and the noise is \( 1 + \lambda + \lambda \delta \) (accounting for regularization and dimensionality effects). When the signal dominates the noise, \( \mathfrak{m}_1 \) increaseses, signifying strong class discrimination, while noise dominance results in \( \mathfrak{m}_1 \to 0 \), indicating poor discrimination. This framework captures the interplay between signal strength, regularization, and dimensionality effects.

\subsection{Analysis of the variance of the decision score}
\label{sec:analysis_variance}
As performed previously, we will experiment also the variation of the variance of the decision score. The variance term can be splitted into four terms.

The first term of the variance is rigorously defined as
\begin{equation*}
    \mathcal{V}_1 = \frac{1}{n^2 m}\tilde{\vy}^\top \mV \tilde{\vy}
\end{equation*}
where $\mV$ is defined as the diagonal matrix containing on its diagonal the vector $\vv = [\frac 1n \tr(\mC_1 \mQ\mC_1\mQ)\mathds{1}_{n_1}, \frac 1n \tr(\mC_2 \mQ\mC_2\mQ)\mathds{1}_{n_2}]$. The term can be rewritten in the more convenient manner as
\begin{equation*}
    \mathcal{V}_1 = \tilde{\vy}^\top\mathcal{D}_\vc \tilde{\vv} \mathcal{D}_\vc\tilde{\vy}
\end{equation*}
Using the covariance matrix identity assumption, one can further get
\begin{equation*}
    \mathcal{V}_1 = \frac{c_0}{(1+\lambda +\lambda\delta)^2 - c_0}
\end{equation*}

The variance of \( g(x) \) is composed of four terms:
\[
\text{Var}(g(x)) = \text{term}_1 + \text{term}_2 + \text{term}_3 + \text{term}_4,
\]
where:
\begin{align*}
\text{term}_1 &= \frac{c_0}{\left((1+\lambda+\lambda\delta)^2 - c_0\right)m}, \\
\text{term}_2 &= \frac{1}{m} \left( A_1 \vmu^\top\vmu + A_2 (\vmu^\top\vmu)^2 + A_3 (\vmu^\top\vmu)^3 + A_4 (\vmu^\top\vmu)^4 \right), \\
\text{term}_3 &= -\frac{2}{m} \frac{\Delta_0 \left(\kappa \vmu^\top\vmu - \gamma (\vmu^\top\vmu)^2\right)}{(1+\delta)^3}, \\
\text{term}_4 &= \frac{m(m-1)}{m^2} \frac{\kappa^2 \vmu^\top\vmu - 2\gamma\kappa (\vmu^\top\vmu)^2 + \gamma^2 (\vmu^\top\vmu)^3}{(1+\delta)^2}.
\end{align*}
We will simplify each term step by step, and then derive the intuition.

\section*{Step 1: Simplify Each Term}

\subsection*{Term 1:}
\[
\text{term}_1 = \frac{c_0}{\left((1+\lambda+\lambda\delta)^2 - c_0\right)m}.
\]
This term reflects the impact of regularization (\( \lambda \)) and the effective dimensionality (\( \delta \)). For large \( m \), this term diminishes due to the \( 1/m \) factor.

\subsection*{Term 2:}
\[
\text{term}_2 = \frac{1}{m} \left( A_1 \vmu^\top\vmu + A_2 (\vmu^\top\vmu)^2 + A_3 (\vmu^\top\vmu)^3 + A_4 (\vmu^\top\vmu)^4 \right),
\]
where:
\[
A_1 = \frac{\kappa^2 S}{(1+\delta)^2}, \quad
A_2 = -\frac{2\gamma\kappa S}{(1+\delta)^2} + \frac{\kappa^2 d_\text{test}}{(1+\delta)^4},
\]
\[
A_3 = \frac{\gamma^2 S}{(1+\delta)^2} - \frac{2\kappa\gamma d_\text{test}}{(1+\delta)^4}, \quad
A_4 = \frac{\gamma^2 d_\text{test}}{(1+\delta)^4}.
\]
Here, \( S = 1 + \frac{d_\text{test}}{(1+\delta)^2} \) and $d_\text{test} = \frac{c_0\kappa^2(1+\delta)^2}{(1+\delta)^2 - c_0\kappa^2}$ where $\kappa = \frac{1}{\frac{1}{1+\delta} + \lambda}$ and $\gamma = \frac{\kappa^2}{(1+\delta)+ \kappa\vmu^\top\vmu}$. Each term in \( A_1, A_2, A_3, A_4 \) decays with increasing \( \delta \) or \( \lambda \), reducing the influence of high-order terms in \( \vmu^\top\vmu \).

\subsection*{Term 3:}
\[
\text{term}_3 = -\frac{2}{m} \frac{\Delta_0 \left(\kappa \vmu^\top\vmu - \gamma (\vmu^\top\vmu)^2\right)}{(1+\delta)^3}.
\]
This term is dominated by \( \Delta_0 \) and decreases inversely with \( m \). For large \( \delta \), this term becomes negligible due to the \( (1+\delta)^{-3} \) scaling.

\subsection*{Term 4:}
\[
\text{term}_4 = \frac{m(m-1)}{m^2} \frac{\kappa^2 \vmu^\top\vmu - 2\gamma\kappa (\vmu^\top\vmu)^2 + \gamma^2 (\vmu^\top\vmu)^3}{(1+\delta)^2}.
\]
As \( m \to \infty \), the factor \( \frac{m(m-1)}{m^2} \to 1 \). This term captures the interaction between subsets, primarily governed by \( \kappa \) and \( \gamma \). For large \( \delta \), it decays due to \( (1+\delta)^{-2} \).

\section*{Step 2: Combine Terms and Simplify}

Combining all terms yields:

\begin{align*}
    \text{Var}(g(x)) 
= & \frac{1}{m} \left[ c_0 \frac{1}{(1+\lambda+\lambda\delta)^2 - c_0} + \sum_{k=1}^4 A_k (\vmu^\top\vmu)^k \right] \\
    & - \frac{2}{m} \frac{\Delta_0 \left(\kappa \vmu^\top\vmu - \gamma (\vmu^\top\vmu)^2\right)}{(1+\delta)^3}
        + \frac{m-1}{m} \frac{\kappa^2 \vmu^\top\vmu - 2\gamma\kappa (\vmu^\top\vmu)^2 + \gamma^2 (\vmu^\top\vmu)^3}{(1+\delta)^2}.
\end{align*}

For large \( m \), terms proportional to \( 1/m \) dominate initially, while for small \( m \), the interaction term (\( \text{term}_4 \)) becomes significant.

\section*{Step 3: Derive Intuition}

\begin{itemize}
    \item \textbf{Dependence on \( m \):} The variance decreases with \( m \) due to the \( 1/m \) scaling of terms like \( \text{term}_1, \text{term}_2, \text{term}_3 \). However, the interaction term (\( \text{term}_4 \)) introduces diminishing returns for large \( m \).

    \item \textbf{Role of Regularization (\( \lambda \)):} Higher \( \lambda \) stabilizes the variance by increasing the denominator of \( \text{term}_1 \) and reducing the magnitudes of \( A_1, A_2, A_3, A_4 \). This effect is particularly strong for weak signals (\( \vmu^\top\vmu \to 0 \)).

    \item \textbf{Dimensionality Effect (\( \delta \)):} As \( \delta \to \infty \), terms like \( (1+\delta)^{-k} \) suppress the variance, reflecting the curse of dimensionality. However, this decay can also mask the signal for large \( \vmu^\top\vmu \).

    \item \textbf{Signal Strength (\( \vmu^\top\vmu \)):} Strong signal (\( \vmu^\top\vmu \to \infty \)) amplifies higher-order terms in \( \text{term}_2 \) and \( \text{term}_4 \), emphasizing the model's ability to discriminate between classes. Weak signals are dominated by regularization and noise.
\end{itemize}

The variance \( \text{Var}(g(x)) \) depends intricately on the interplay between \( m \), \( \lambda \), \( \delta \), and \( \vmu^\top\vmu \). Increasing \( m \) reduces variance initially but exhibits diminishing returns due to the interaction term. Strong regularization (\( \lambda \)) and high dimensionality (\( \delta \)) stabilize the variance, though they may suppress meaningful signal (\( \vmu^\top\vmu \)).

\end{document}